\documentclass{article}

 \usepackage[main, final]{neurips_2026}

\usepackage[utf8]{inputenc} 
\usepackage[T1]{fontenc}    
\usepackage{hyperref}       
\usepackage{url}            
\usepackage{booktabs}       
\usepackage{amsfonts}       
\usepackage{nicefrac}       
\usepackage{microtype}      
\usepackage{xcolor}         
\usepackage{pgf}
\usepackage{booktabs}
\usepackage{multirow}
\usepackage{makecell}
\usepackage{thmtools}
\usepackage{siunitx}

\usepackage{amsmath,amsfonts,bm}

\def\eqref#1{equation~\ref{#1}}

\def\1{\bm{1}}

\def\vt{{\bm{t}}}

\def\vx{{\bm{x}}}

\def\vz{{\bm{z}}}

\def\vdelta{{\bm{\delta}}}

\def\mI{{\bm{I}}}

\DeclareMathAlphabet{\mathsfit}{\encodingdefault}{\sfdefault}{m}{sl}
\SetMathAlphabet{\mathsfit}{bold}{\encodingdefault}{\sfdefault}{bx}{n}

\def\gB{{\mathcal{B}}}

\def\gD{{\mathcal{D}}}
\def\gE{{\mathcal{E}}}

\def\gL{{\mathcal{L}}}

\def\gN{{\mathcal{N}}}

\def\gX{{\mathcal{X}}}
\def\gY{{\mathcal{Y}}}

\def\sI{{\mathbb{I}}}

\newcommand{\E}{\mathbb{E}}

\DeclareMathOperator*{\argmax}{arg\,max}
\DeclareMathOperator*{\argmin}{arg\,min}

\newcommand{\vepsilon}{\boldsymbol{\epsilon}}
\newcommand{\setIterate}[3]{\{ #3 \}_{#1}^{#2}}
\usepackage{amsmath}
\usepackage{amssymb}
\usepackage{amsthm}
\usepackage{mathtools}
\usepackage{xspace}
\usepackage{fontawesome5}
\usepackage{algorithm}
\usepackage{algpseudocode}

\definecolor{proofdarkred}{RGB}{140,0,0}
\definecolor{proofdarkgreen}{RGB}{0,100,35}

\newif\ifreview
\reviewtrue
\definecolor{reviewremoved}{RGB}{210,0,0}
\definecolor{reviewadded}{RGB}{0,140,40}
\definecolor{reviewnotecolor}{RGB}{230,120,0}
\ifreview
  
  \DeclareRobustCommand{\chgnote}[1]{{\color{reviewnotecolor}[#1]}}
\else
  
  \DeclareRobustCommand{\chgnote}[1]{}
\fi

\newtheorem{proposition}{Proposition}

\usepackage[inline]{enumitem}
\usepackage{pgfplots}

\renewcommand{\arraystretch}{0.85}

\newcommand{\mathset}[1]{\left\{ #1 \right\}}

\newcommand{\defeq}{\vcentcolon=}

\newcommand{\Dplus}{\gD}

\newcommand{\dcal}{\gD_{n}}
\newcommand{\xtest}{\vx_{n+1}}
\newcommand{\xtestpert}{\tilde\vx_{n+1}}
\newcommand{\lowerval}{\mathrm{c}^\downarrow}
\newcommand{\upperval}{\mathrm{c}^\uparrow}

\makeatletter
\newcommand{\figpartaux}[2]{%
  \global\@namedef{figpart@#1}{#2}%
}
\newcommand{\figpartlabel}[2]{%
  \begingroup
  \def\@currentlabel{\thefigure}%
  \def\@currentHref{figure.\theHfigure}%
  \label{#1}%
  \protected@write\@auxout{}{\string\figpartaux{#1}{#2}}%
  \endgroup
}
\newcommand{\figpartname}[1]{%
  \@ifundefined{figpart@#1}{??}{\@nameuse{figpart@#1}}%
}
\newcommand{\subautoref}[1]{\hyperref[#1]{\autoref*{#1} - \figpartname{#1}}}
\makeatother
\newcommand{\cert}[1]{\mathrm{cert}_{#1}}
\newcommand{\proxy}[1]{\hat{\mathrm{cert}}_{#1}}

\newcommand{\certproxy}{\hat{\mathrm{cert}}}
\newcommand{\method}{ProxC\xspace}

\title{Should We Pay This Much for Robustness? \\ Efficient Proxy Certificates with Marginal Guarantees}

\author{%
  Soroush H. Zargarbashi$^{\star, 1}$, Mohammad Sadegh Akhondzadeh$^{\star, 2}$, Aleksandar Bojchevski$^{2}$ \\
  $^1$CISPA Helmholtz Center for Information Security, $^2$University of Cologne\\
  \texttt{[zargarbashi, akhondzadeh, bojchevski]@cs.uni-koeln.de} \\
}

\begin{document}

\maketitle

\begin{abstract}
Modern classifiers remain sensitive to small input perturbations, while strong model-agnostic robustness certificates such as randomized smoothing are often too expensive for deployment, requiring an extensive number (e.g. $2000$) of model forward passes per input. We propose a framework that delegates certification to a cheap proxy function while statistically controlling its deviation from the original sound certificate. Given an exchangeable unlabeled calibration set, we tune the proxy via robust conformal risk control to guarantee a bounded marginal (rather than pointwise) false-approval rate: the probability, over future inputs, that the proxy accepts an input whose prediction is flipped due to perturbation. The resulting proxy certificate can be used either as a standalone marginal certificate or as a fallback mechanism that invokes the expensive certificate only when the proxy rejects. We provide both single-sample and sample-heavy calibration recipes. Empirically, as a standalone certificate our method reduces the test-time certification cost to a single model forward while retaining much of the certified performance of randomized smoothing at small radii, and as a fallback it recovers the certified ratio of randomized smoothing at a more modest speedup. This enables cheap yet effective robustness guarantees for larger architectures such as vision transformers and language models. 
\end{abstract}

\section{Introduction}
\label{sec:introduction}

High accuracy alone is not sufficient for reliability in machine learning. Although modern neural networks achieve strong empirical performance, their predictions can be highly sensitive to small, often imperceptible, input perturbations \citep{goodfellow2014explaining}. This sensitivity can be exploited by adversaries to easily degrade the model's performance. There are many empirical, and heuristic defenses proposed in the literature which have been circumvented by stronger or adaptive attacks \citep{athalye2018obfuscated}. A principled way to address this issue is to use robustness certificates, which provide formal guarantees that the model prediction remains invariant within a predefined perturbation set. Certificates can be viewed as binary decision rules: when they certify an input at a sufficiently large radius, they signal that the prediction can be trusted with respect to the specified threat model, whether the input is clean or even if adversarially perturbed, the perturbation is not effective.

\begin{table}[t]
\centering
\caption{Results on ImageNet/ResNet, $r{=}0.14$, $\sigma{=}0.5$,
$\alpha{=}0.05$, Gaussian smoothing ($2000$ MC samples: $50$ for the majority-vote label and $1950$ for the certificate). The verification entry does not use smoothing (result directly reported from the reference); it gives a pointwise guarantee on a different architecture, so it is not directly comparable to the marginal guarantee of \method.
Speedup is on forward calls and relative to randomized smoothing. Certification-only cost excludes the forwards for the label; end-to-end cost includes them.}
\label{tab:timing}
\setlength{\tabcolsep}{3pt}
\small
\resizebox{\linewidth}{!}{%
\begin{tabular}{lllcccccc}
\toprule
\multirow{2}{*}{Method} & \multirow{2}{*}{Variant} &
\multirow{2}{*}{Guarantee} &
\multirow{2}{*}{\makecell{Cert.\\Ratio}} &
\multirow{2}{*}{\makecell{Cert.\\Accuracy}} &
\multirow{2}{*}{\makecell{Error\\Rate}} &
\multicolumn{3}{c}{Forwards / image} \\
\cmidrule(l){7-9}
& & & & & & Cal & \makecell{Cert.-only\\{\footnotesize(Speedup)}} & \makecell{End-to-end\\{\footnotesize(Speedup)}} \\
\midrule
\addlinespace[4pt]
Rand.\ Smoothing & ---
& Pointwise
& 70.06\% & 56.08\% & ---
& --- & 1950 & 2000 \\
\addlinespace[4pt]
Verification & \citet{hu2023unlocking}
& Pointwise
& --- & 35.0\% & ---
& --- & 1 & 1 \\
\addlinespace[2pt]
\midrule
\addlinespace[4pt]
\multirow{2}{*}{ProxC}
& Standalone
& Marginal
& 56.11\% & 48.53\% & 2.29\%
& 2000 & 1 {\footnotesize $\mathbf{(1950\times)}$} & 51 {\footnotesize $\mathbf{(39.2\times)}$} \\
& Fallback
& Marginal
& 72.35\% & 57.3\% & 2.29\%
& 2000 & 779 {\footnotesize $\mathbf{(2.50\times)}$} & 829 {\footnotesize $\mathbf{(2.41\times)}$} \\
\addlinespace[2pt]
\bottomrule
\end{tabular}}
\end{table}

Existing robustness certificates typically obtain such guarantees through architecture-specific verification methods, such as convex relaxations, or mixed-integer programming \citep{wong2018convex,katz2017reluplex,tjeng2019evaluating}. While powerful, these methods require white-box access to the model, and often need to be adapted when new architectures or layers are introduced. Alternatively, randomized smoothing (RS) \citep{cohen2019certified} wraps around any black-box classifier, and convolves it with a noise augmentation. The resulting smoothed classifier changes slowly with local perturbations. Hence, one can derive closed-form model-agnostic bounds on its change. This black-box nature is a major practical advantage, allowing it to easily apply to large modern networks, and even to API-accessible systems \citep{carlini2022certified}. However, randomized smoothing-based certificates require a large number of model forward passes (e.g. $2 \times 10^3$) to estimate the smoothed class probabilities with high confidence, which can be prohibitively expensive especially with limited resources, or when the model itself is expensive to evaluate. 

This cost is especially wasteful because not every input received by a system is adversarial; indeed, most inputs are expected to be clean. Moreover, in many settings, it is cheap to obtain a relatively small set of unlabeled samples representative of the deployment distribution. These observations motivate a distributional relaxation of robustness certification: rather than requiring a local certificate for every individual input, it may suffice to guarantee that certification failure is controllable below an adjustable low probability over test inputs.

In light of the arguments above, we define \underline{Prox}y \underline{C}ertificate (\method), a framework for delegating expensive certification to cheap proxy functions under calibrated statistical control. Through a variant of conformal risk control (CRC) that we define in \autoref{sec:conformalRiskControl}, we control how often the proxy deviates from the expensive function. Our proxy reduces the cost of the certification step to a single model forward. We define three variants for \method, tailored for various compute budgets:\method-A, that uses a single forward pass per calibration point, \method which computes an expensive backbone certificate on the small calibration set, and \method-H which can increase the certified ratio, in return for more Monte-Carlo (MC) sampling. All three variants use a single sample to certify each test point. We bound the risk anchored at the latent clean point, which lets a sound certificate computed only on clean calibration points calibrate a cheap proxy that remains valid under perturbation (see \autoref{sec:related}).


The resulting proxy can be used either as a standalone certificate or as a hybrid fallback mechanism that only calls the expensive certificate on inputs rejected by the proxy. These are two different operating regimes: the standalone proxy gives the largest saving, while the fallback recovers the certified ratio of the backbone at a smaller speedup.
As we show in \autoref{tab:timing}, by relaxing the soundness to bounded false approvals, we can reduce the cost of the certification step to a single forward, the same as verification-based approaches (importantly, the choice of label passed into the certificate is excluded from the cost evaluation. This can still follow the inference rule of the backbone). \method shows a higher certified accuracy compared to verification (note that our guarantee is marginal, and we use a slightly different architecture), and it is cheaper than randomized smoothing. In other words, in the trade-off between computational efficiency and applicability, we now propose a new dimension which is soundness. We propose smoothing-based guarantees with a realistic computational cost that is suitable for larger models like ViT, or language modeling in realistic time-sensitive scenarios.

\section{Background}
\label{sec:background}

\paragraph{Problem setup.}
Let $f: \mathcal{X} \to \mathbb{R}^{|\mathcal{Y}|}$ be a pre-trained classifier (can be a single model, a pipeline of models, or any general inference policy, e.g. majority vote over random augmentations) that returns a score per class. For an input $\vx_i \in \mathcal{X}$, its top prediction is $y_{i} = \argmax_y f(\vx_i)_y$. For brevity, whenever $f$ is compared across two inputs, e.g. $f(\tilde\vx) \neq f(\vx)$, we compare the predicted labels, i.e. $\argmax_y f(\tilde\vx)_y \neq \argmax_y f(\vx)_y$. For the test point, $\xtest$, an adversary introduces a perturbation $\xtestpert = \xtest + \vdelta$ aiming to change the model's prediction $f(\xtestpert) \neq f(\xtest)$. The adversary is limited to keep the perturbation imperceptible -- $\xtestpert$ has to be very close to $\xtest$. This limitation is formally presented as a perturbation ball $\gB: \gX \to 2^\gX$ -- the set of all points from a given $\vx$ that the adversary is allowed to choose. A common example for images is the $\ell_2$ ball $\gB_r(\vx) = \{\tilde\vx : \|\tilde\vx - \vx\|_2 \le r\}$. The adversary has white-box access to the model (its architecture, parameters, etc), and the clean input $\xtest$, but is not in control of the inference, and is unaware of the random seed used by the MC samples evaluated at test time. The test-time noise is freshly sampled for every query and is unknown when the adversary chooses its perturbation; hence we do not require a permanently secret global seed. We study the \emph{worst-case} robustness: for \emph{any} $\xtestpert \in \gB_r(\xtest)$, the prediction should either remain the same, or the defense should abstain from certifying it. This is what we call the threat model. 

\paragraph{Robustness certificate.} A robustness certificate is a binary decision function
\begin{align}
    \label{eq:robustness-certificate}
    \cert{f,r}(\vx_i) = 1 \qquad \Rightarrow \qquad \forall \tilde\vx \in \gB_r(\vx_i) : f(\tilde\vx) = f(\vx_i).
\end{align}
The certificate is \emph{sound}: $\cert{f,r}(\vx_i) = 1$ ensures $f$ is stable entirely within $\gB_r(\vx_i)$, but it is not \emph{complete}: $\cert{f,r}(\vx_i) = 0$ does not necessarily imply vulnerability. The standard use case of the certificate is to apply it at test time on each (potentially perturbed) input $\xtestpert$, and if $\cert{f,r}(\xtestpert) = 1$, we can safely use the prediction $f(\xtestpert)$; otherwise, we abstain. 

\paragraph{Inverse ball $\gB^{-1}$.} While we allow the adversary to perturb $\vx$ within $\gB(\vx)$, from the defender's perspective, we should ensure that the received $\tilde\vx$ has the same prediction as $\vx$ which is the reverse act w.r.t. the adversary. Hence, at the test time we use the certificate over the inverse ball $\gB^{-1}(\vx) = \mathset{\tilde\vx: \vx \in \gB(\tilde\vx)}$ which is the smallest set that from any perturbed point $\tilde\vx \in \gB(\vx)$ it includes $\vx$. This is the ball that covers the clean point from any potential perturbation. In symmetric balls, we have $\gB^{-1}(\vx) = \gB(\vx)$. An example for non-symmetric ball is token insertion, or deletion, where the inverse ball is the opposite act; i.e. $\gB_{r}^\mathrm{ins, -1} =\gB_{r}^\mathrm{del}$.

\paragraph{Binary and confidence certificate.} We use $\lowerval$ and $\upperval$ for certified lower and upper bounds, with a slight overload: formally $\lowerval(f, \gB(\vx)) \le \inf_{\tilde\vx\in\gB(\vx)} f(\tilde\vx)$ and $\upperval$ is the corresponding upper bound defined with supremum. In case that the bounds only depend on the value $p=f(\vx)$ (not the function, and the point), we write $\lowerval(p, \gB)$ and $\upperval(p, \gB)$. Certificates often report $\lowerval$, and $\upperval$, from which we obtain the 0/1 decision or the certified radii -- the largest radius for which the prediction is provably invariant; i.e. $r^*_{\vx} = \max \{ r : \lowerval(f, \gB_r(\vx)) \ge 1/2 \}$. The latter is a common presentation of certificates; larger radii indicate stronger robustness. However ultimately the certification of a point boils down to comparison with a predefined radius $r$, which is a binary decision.
We say $\vx$ is certified at radius $r$ either if
$
\lowerval(p_{y^\star}(\vx), \gB_r) > 1/2
$
or if
$
\lowerval(p_{y^\star}(\vx), \gB_r) > \max_{y \neq y^\star} \upperval(p_y(\vx), \gB_r)
$, for $y^\star$ as the top predicted class.
In this work, we operate directly on these certified bounds.

Common certificates are often obtained via exact verification methods \citep{wong2018convex,katz2017reluplex} and randomized smoothing \citep{cohen2019certified, yang2020randomized}. The former is a white-box method that requires access to the model's architecture and parameters, and is typically limited to smaller models of known architectures. The latter is a black-box method that can be applied to any model, and can scale to large and modern architectures, however it requires a large number of model forward passes to estimate the smoothed class probabilities with high confidence.

\paragraph{Objective.}
We replace the sound but expensive certificate with a tuned proxy function that controls the marginal false-approval rate at level $\alpha$. The guarantee holds for all adversarial perturbations $\xtestpert \in \gB(\xtest)$ for any given clean point $\xtest$, and the expectation in \autoref{eq:goal} is taken marginally over the distribution of (latent) clean inputs.
We assume a calibration dataset $\dcal = \{\vx_1, \dots, \vx_n\}$ of unlabeled points drawn exchangeably from the same distribution as the clean test input $\xtest$ over which we tune the certificate. Only the latent clean point $\xtest$, and not $\xtestpert$, needs to be exchangeable with $\dcal$; e.g.\ for unseen deployment inputs, $\dcal$ must be held out from the training data of $f$, as training and test points are not exchangeable conditional on the model. The outcome is a proxy that takes significantly lower computational cost only by relaxing the soundness of the original certificate to a marginally bounded false-approval rate; formally with $\proxy{f, r}$, we guarantee that
\begin{align}
    \label{eq:goal}
    \mathbb{E}_{\mathcal{D}}\!\left[
        \sI\{f(\xtestpert) \neq f(\xtest)\} \cdot \proxy{f, r}(\xtestpert) \mid \xtestpert \in \gB(\xtest)
    \right] \le \alpha.
\end{align}
In words, over many future (possibly perturbed) inputs, on average at most an $\alpha$ fraction are approved while their prediction is flipped. The expectation is over the calibration set, the clean test point, and the internal randomness. Same as conformal prediction itself, the risk is bounded in expectation, and for a fixed calibration set can exceed $\alpha$ (with a variance decreasing in the number of calibration points $n$); the guarantee is also not conditional on any particular input, class, or subpopulation.

Our proxy certificate (\method) is built upon two key components, which we introduce next.

\subsection{Randomized Smoothing}
We recall a high level understanding of randomized smoothing, and defer the details to \autoref{sec:intro-to-rs}. A smoothing function $\psi:\gX\to\gX$ augments the input with a random noise from a predefined distribution; $\psi(\vx) = \vx + \vepsilon$. 
Wrapping any base classifier $f$, we define the smoothed class probability $g(\vx)_y = \Pr_{\vepsilon}[f(\vx + \vepsilon) = y]$. Because the push-forward measures $f_\#\psi(\vx)$ and $f_\#\psi(\tilde\vx)$ overlap heavily for nearby inputs, we can derive model-agnostic lower and upper bounds on $g(\tilde\vx)_y$ that depend only on $p = g(\vx)_y$ and the perturbation radius.
On images a well-known smoothing scheme is the isotropic Gaussian noise of magnitude $\sigma$, i.e. $\vepsilon \sim \gN(0, \sigma^2\mI)$. For $\ell_1$, and $\ell_2$ ball, isotropic Gaussian smoothing has closed form lower and upper bounds ($\lowerval$, and $\upperval$) as $\Phi_\sigma(\Phi^{-1}_\sigma(p) \pm r)$ regardless of the model's architecture, or the actual value of $\vx$. Here $\Phi_\sigma$ is the CDF of $\gN(\boldsymbol{0}, \sigma^2\mI)$. For other smoothing schemes, similar bounds can be computed through the recipe from \cite{yang2020randomized}.

\subsection{Conformal Risk Control}
\label{sec:conformalRiskControl}


CRC provides distribution-free, finite-sample guarantees on the expected risk of a system using an exchangeable subsample of its inputs \citep{angelopoulos2022conformal}. Let $L:\gX\times[0, 1]\to\mathbb{R}$ be a risk function with conservativeness parameter $\lambda$, so that $L(\vx, \lambda)$ is non-increasing in $\lambda$, right continuous, upper bounded by $l_\mathrm{max}$, and $\min_\lambda L(\cdot, \lambda) \le \alpha$ for some user-specified $\alpha$. CRC guarantees that we can find $\lambda^\star$ such that for future test points $\xtest$ drawn from the same distribution, $\E[L(\xtest, \lambda^\star)] \le \alpha$. Over the holdout calibration points $\dcal = \setIterate{i=1}{n}{\vx_i}$, and $\bar L_n(\lambda)=\frac{1}{n}\sum_{i=1}^n L(\vx_i, \lambda)$, under exchangeability of $\Dplus = \dcal \cup \{\xtest\}$ CRC sets
\[
\lambda^\star=\inf\left\{\lambda:\frac{n}{n+1}\bar L_n(\lambda)+\frac{l_\mathrm{max}}{n+1}\le \alpha\right\}, \quad \text{ensuring} \quad \E[L(\xtest, \lambda^\star)] \le \alpha
\]

\section{ProxC: Proxy Certificate}
\label{sec:proxy-cert}

Our proxy certificate is based on an internal binary proxy function $h: \gX \times [0, 1] \to \{0, 1\}$; the second parameter of this function, $\lambda$, controls its conservativeness. Through conformal risk control we compute the smallest valid parameter $\lambda^*$ that guarantees to control the rate of false-approval made by $h(\cdot, \lambda^*)$. 
A direct application of CRC would treat false approval as an ordinary risk, guaranteeing validity of $h(\cdot,\lambda)$ for clean test points that are in-distribution with $\dcal$. However, we aim to maintain the guarantee for any $\xtestpert \in \gB(\xtest)$, not just the clean point. Thus, the risk observed at test time is not exchangeable with the risks used for calibration. We address this mismatch in \autoref{sec:robust-crc}. After obtaining a valid robustly calibrated parameter $\lambda^\star$, we define
$
\proxy{f,r}(\xtestpert)=h(\xtestpert,\lambda^\star),
$
and guarantee
$
\E_{\Dplus}\left[\sI\mathset{f(\xtestpert)\neq f(\xtest)}\cdot h(\xtestpert,\lambda^\star)\right]\le \alpha
$ for all $\xtestpert \in \gB(\xtest)$.

\subsection{Robust Conformal Risk Control Through Randomized Smoothing}
\label{sec:robust-crc}

\citet{angelopoulos2022conformal} propose robust risk control by replacing the pointwise risk $L(\xtest,\lambda)$ with its worst-case local value $\widetilde L(\vx,\lambda)=\sup_{\tilde\vx\in\gB_r(\vx)} L(\tilde\vx,\lambda)$. Calibrating over $\widetilde L$ ensures
\[
\mathbb{E}_{\Dplus}\!\left[
L(\xtestpert,\lambda)\mid \xtestpert\in\gB(\vx)
\right]
\le
\mathbb{E}_{\Dplus}\!\left[
\sup_{\xtestpert\in\gB_r(\xtest)} L(\xtestpert,\lambda)
\right]
\le \alpha.
\]
In practice, however, evaluating $\widetilde L$ is itself a certification problem. While it can be upper bounded via $\upperval(L(\cdot,\lambda),\gB(\vx))$, this reintroduces the computational cost we seek to avoid, as we use the certificate again. 
We instead rewrite the approach of \citet{Zargarbashi2025OneSI} for general risk control, and generalize it to sample-heavy calibration. We apply randomized smoothing at the level of the expected risk. We define a noise-augmented set $\gD^+ = \mathset{\vx + \vepsilon: \vx \in \Dplus}$ that preserves exchangeability, yielding a conformal guarantee that is marginal over both the data distribution and the smoothing noises. Here, randomized smoothing bounds transfer this guarantee (conditional on each individual $\xtest$) to all perturbations in $\gB_r(\xtest)$, enabling robust single-sample inference without Monte Carlo estimation at deployment.

\begin{restatable}{proposition}{robustrisk}
    \label{thrm:robust-rs-risk}
    Let $\dcal = \mathset{\vx_1,\dots,\vx_n}$ be exchangeable together with the (potentially unseen) clean test point $\xtest$.  Let $L:\gX\times[0,1]\to\{0,1\}$ be such that, for every $\vx \in \gX$, the map $\lambda\mapsto L(\vx,\lambda)$ is non-increasing, right-continuous, and we have $L(\cdot, 1) = 0$. Let $\psi:\gX\to\gX$ be a smoothing scheme that is independent of the data, with model-agnostic certified bounds $\lowerval(\cdot,\gB^{-1})$ and $\upperval(\cdot,\gB)$. Given any user-specified tolerance $\alpha > 0$, let $\vepsilon_1,\ldots,\vepsilon_{n+1}$ be iid draws from $\psi$. With $\lambda^*$ defined as either
    \begin{align}
        \label{eq:lambda-cert-single}
        \lambda^* = \inf\{\lambda: \frac{1}{n+1} \left[ 1 + \sum_{i = 1}^{n} L(\vx_i + \vepsilon, \lambda)\right] \le \lowerval(\alpha, \gB^{-1}) \}
    \\
    \text{or} \quad 
        \label{eq:lambda-cert-heavy}
        \lambda^* = \inf\{\lambda: \frac{1}{n+1} \left[ 1 + \sum_{i = 1}^{n} \upperval_e(\Pr_{\vepsilon}[L(\vx_i + \vepsilon, \lambda) = 1], \gB) \right] \le \alpha - \delta \}
    \end{align} where the latter $\lambda^*$ corresponds to the calibration-heavy variant,
    we have
    \[
    \mathbb{E}_{\Dplus, \vepsilon}\!\left[L(\xtestpert + \vepsilon, \lambda^*)\right] \le \alpha, \quad \forall \xtestpert \in \gB(\xtest).
    \]
    where $\delta \in [0, 1]$ is the confidence of the finite sample estimation, and $\upperval_e(\Pr_{\vepsilon}[L(\vx + \vepsilon, \lambda) = 1], \gB)$ is an upper confidence bound on $\upperval(\Pr_{\vepsilon}[L(\vx + \vepsilon, \lambda) = 1], \gB)$ with confidence $1 - \frac{\delta}{n}$.
\end{restatable}

Proofs are in \autoref{sec:proofs}. For the calibration-heavy approach, we estimate the risk at each point $\vx_i$ by averaging over $m$ noises $\{\vepsilon_{i,j}\}_{j=1}^m$; i.e. $\hat{p}_i(\lambda) = \frac{1}{m} \sum_{j=1}^m L(\vx_i + \vepsilon_{i,j}, \lambda)$. We then construct the Clopper-Pearson upper confidence bound $\overline{p}_i(\lambda)$ with confidence $1 - \delta/n$, and set $\upperval_e\!\bigl(\Pr_{\vepsilon}[L(\vx_i + \vepsilon, \lambda) = 1], \gB\bigr) \coloneqq \upperval(\overline{p}_i(\lambda), \gB)$. Since $\upperval$ is monotone and $\overline{p}_i(\lambda) \ge \Pr_{\vepsilon}[L(\vx_i + \vepsilon,\lambda)=1]$ with the stated confidence, this construction is a valid upper bound on the true smoothed risk (see \autoref{sec:flor-proxch}).

\subsection{\method: Conformal Proxy Certificate}
\label{sec:ProxC}

Let $\hat{h}(\cdot, \lambda)$ be a binary proxy function that is controlled by a conservativeness parameter $\lambda$ -- for any $\vx \in \gX$ and $\lambda_1 < \lambda_2$, we have $\hat{h}(\vx, \lambda_1) \ge \hat{h}(\vx, \lambda_2)$; a trivial example for proxy function is $\hat{h}(\vx, \lambda) = \sI[\max_{y} f(\vx) > \lambda]$. We expect the certificate, and the proxy function to ensure that $f(\xtestpert) = f(\xtest)$; hence, we directly define the false approval risk for $\hat{h}$ as
\[
    \Pr_{\Dplus}\!\left[
        \sI\{f(\xtestpert) \neq f(\xtest)\} \cdot \hat{h}(\xtestpert, \lambda^*)\right] \le \alpha \quad \forall \xtestpert \in \gB(\xtest).
\]
From \autoref{thrm:robust-rs-risk}, we concluded that augmenting $\vx$ with a random noise results in black-box extension of the risk control guarantee to the whole $\gB(\xtest)$. The risk defined above does not allow the proxy function for any inherent randomness. Hence, we replace the deterministic risk with the following probabilistic generalization 
\begin{align}
    \label{eq:proxy-risk-solid}
    \Pr_{\Dplus, \vepsilon}\!\left[
        \sI\{f(\xtestpert) \neq f(\xtest)\} \cdot \hat{h}(\xtestpert + \vepsilon, \lambda^*)\right] \le \alpha \quad \forall \xtestpert \in \gB(\xtest) \nonumber\\
    \text{equivalently} \quad \E_{\Dplus}\left[ \max_{\xtestpert\in\gB(\xtest)}\sI\{f(\xtestpert) \neq f(\xtest)\} \cdot \Pr_{\vepsilon}\!\left[\hat{h}(\xtestpert + \vepsilon, \lambda^*) = 1\right]\right] \le \alpha.
\end{align}
This also generalizes the threat model. In a deterministic setup, the adversary's goal is to change the model's prediction on an input. In case the certificate was not sound, a more realistic goal was to change the prediction, and fool the certificate to certify the point.
Here, with the probabilistic notion of certificate, the goal of the adversary is to increase the probability of the proxy to approve a point while the label is also flipped. 
We guarantee the above risk to be bounded through the following:

\begin{proposition}
    \label{thrm:proxy-cert}
    Let the proxy function $\hat{h}$ be non-increasing and right continuous in $\lambda$, and let $\hat{h}(\cdot, 1) = 0$. Let $\cert{f, r}$ be a sound certificate with respect to $\gB_r(\cdot)$, and let $\psi$ be a smoothing scheme with model-agnostic upper, and lower bounds $\upperval$, and $\lowerval$. Then the proxy function $\hat{h}(
    \cdot+ \vepsilon, \lambda^*)$ with either
    \begin{align}
    \label{eq:proxy-crc-lambda-cheap}
    \lambda^* = \inf\left\{ \lambda: \frac{1}{n+1}\left[1 + \sum_{i = 1}^n \hat{h}(\vx_i + \vepsilon, \lambda) \cdot (1 - \cert{f, r}(\vx_i))\right] \le \lowerval(\alpha, \gB^{-1}) \right\}\\
    \label{eq:proxy-crc-lambda-heavy}
    \text{or} \qquad \lambda^* = \inf\left\{ \lambda: \frac{1}{n+1}\left[ 1 + \sum_{i = 1}^n \upperval(\hat{h}_\lambda, \gB(\vx_i)) \cdot (1 - \cert{f, r}(\vx_i))\right] \le \alpha \right\}
    \end{align}
    guarantees \autoref{eq:proxy-risk-solid} to hold.
\end{proposition}

\begin{proof}
    Given that the certificate is sound, for any $\vx$, and $\tilde\vx \in \gB(\vx)$, we have $\sI\{ f(\tilde\vx) \neq f(\vx)\} \le 1 - \cert{f, r}(\vx)$, and therefore \begin{align*}
        \max_{\xtestpert\in\gB(\xtest)}\sI\{f(\xtestpert) \neq f(\xtest)\} \cdot \Pr_{\vepsilon}\!\left[\hat{h}(\xtestpert + \vepsilon, \lambda^*) = 1\right] \\
        \le \max_{\xtestpert\in\gB(\xtest)}\sI\{f(\xtestpert) \neq f(\xtest)\} \cdot \max_{\xtestpert\in\gB(\xtest)} \Pr_{\vepsilon}\!\left[\hat{h}(\xtestpert + \vepsilon, \lambda^*) = 1\right] \\
        \le (1 - \cert{f, r}(\xtest)) \cdot \max_{\xtestpert\in\gB(\xtest)}\Pr_{\vepsilon}\!\left[\hat{h}(\xtestpert + \vepsilon, \lambda^*) = 1\right] \\
        \le (1 - \cert{f, r}(\xtest)) \cdot \upperval(\hat{h}_\lambda, \gB(\xtest))
    \end{align*}

    The risk function has two inputs besides $\lambda$ -- the received point and its ground truth clean point, as an anchor. The anchor variable $\xtest$ should be kept explicit.  We define the risk for the exchangeable paired sample $\{(\vx_i,\vx_i)\}_{i=1}^{n}\cup\{(\xtest,\xtest)\}$, and only allow the first coordinate to be perturbed.  For the single-sample rule define the binary anchored risk
    \[
        L_{\mathrm{anc}}((\vz,\vx),\lambda) 
        =
        (1-\cert{f,r}(\vx))\,\hat h(\vz,\lambda).
    \]  
    This risk is non-increasing, right-continuous, and bounded in $\lambda$ as it inherits properties from $\hat{h}$. Note that the other term is only 0, or 1, and only a function of the second element $\vx$. Applying
    \autoref{thrm:robust-rs-risk} in single-sample form, i.e. \autoref{eq:lambda-cert-single}, tunes to $\lambda^*$ as defined in \autoref{eq:proxy-crc-lambda-cheap}, and guarantees that
    \[
        \E_{\Dplus,\vepsilon}\!\left[
        (1-\cert{f,r}(\xtest))\hat h(\xtestpert+\vepsilon,\lambda^*)
        \right]\le\alpha,
        \qquad \forall \xtestpert\in\gB(\xtest).
    \]
    Note that here, the risk control is applied to the tuple, and the certificate only to the first argument (the anchor). The soundness inequality
    $\sI\{f(\xtestpert)\neq f(\xtest)\}\le 1-\cert{f,r}(\xtest)$ then implies
    \[  
        \E_{\Dplus,\vepsilon}\!\left[
        \sI\{f(\xtestpert)\neq f(\xtest)\}\hat h(\xtestpert+\vepsilon,\lambda^*)
        \right]\le\alpha \quad \forall \xtestpert \in \gB(\xtest).
    \]  
    For the calibration-heavy rule, we compute the worst-case smoothed risk of the anchored loss. Since the certificate term $(1-\cert{f,r}(\vx))$ depends only on the anchor $\vx$ and not on the perturbed first argument, $
        \upperval(L_{\mathrm{anc}}((\cdot,\vx),\lambda), \gB(\vx))
        =
        (1-\cert{f,r}(\vx))\,\upperval(\hat h_\lambda,\gB(\vx))
    $.
    Applying the calibration-heavy form of \autoref{thrm:robust-rs-risk} (i.e.\ \autoref{eq:lambda-cert-heavy}) with this upper bound yields \autoref{eq:proxy-crc-lambda-heavy} and the same final guarantee \autoref{eq:proxy-risk-solid}. The core intuition throughout is to define the risk as a function of \emph{both} the perturbed input and the corresponding clean anchor. During calibration, both coordinates coincide and are observable; at test time, the guarantee holds with respect to the certificate evaluated at the (latent) clean point.
\end{proof}

\subsection{Augmented Inference}

\autoref{thrm:proxy-cert} requires an existing sound certificate at calibration time, which is itself computationally expensive. An alternative is to \emph{incorporate} the noise augmentation directly into the inference rule, defining a probabilistic classifier which already exhibits robustness. In this case, the smoothed inference simultaneously produces a prediction and serves as an abstention rule: the proxy approves when the noise-augmented model agrees with the clean prediction, and the guarantee covers the event that the prediction at $\xtestpert + \vepsilon$ toggles from that of $\xtest$, and the proxy nevertheless approves. This is a guarantee on a randomized joint prediction-and-approval event: both the label and the approval are obtained from the same noise draw. The guarantee (for bounded false approval ratio) is written as:
\begin{align}
    \label{eq:proxy-risk-augmented}
    \E_{\Dplus}\left[\max_{\xtestpert \in \gB(\xtest)} \Pr_{ \vepsilon}\!\left[
        \sI\{f(\xtestpert + \vepsilon) \neq f(\xtest)\} \cdot \hat{h}(\xtestpert + \vepsilon, \lambda^*)\right]\right] \le \alpha
\end{align} 

\begin{proposition}
    \label{thrm:proxy-cert-augmented}
    Given the proxy function $\hat{h}$, and the smoothing distribution $\psi$ with the same properties as in \autoref{thrm:proxy-cert}, the proxy function $h(\vz) = \hat{h}(\vz + \vepsilon, \lambda^*)$ with
    \begin{align*}
        \lambda^* = \inf\left\{ \lambda: \frac{1}{n+1}\left[1 + \sum_{i = 1}^n \eta(\vx_i + \vepsilon, \lambda, \vx_i) \right] \le \lowerval(\alpha, \gB^{-1}) \right\}
    \end{align*} and $\eta(\vz, \lambda, \vx) = \hat{h}(\vz, \lambda) \cdot \sI\{f(\vz) \neq f(\vx)\}$ guarantees \autoref{eq:proxy-risk-augmented} to hold.
\end{proposition}

\subsection{\method: The framework}
\label{sec:framework}
While our framework is general, we focus on a trivial example which is a thresholded proxy function $\hat h(\vz,\lambda) = \sI[f(\vz)_y > \lambda]$. Before deployment, we calibrate the proxy on $\dcal$ using noise-augmented inputs, following either \autoref{thrm:proxy-cert} or \autoref{thrm:proxy-cert-augmented} depending on the desired guarantee, and with either a single noise sample per point or a larger MC budget.
Intuitively, for \method we run the expensive certificate once per calibration point; only the points it does not certify can be falsely approved, and we pick the least conservative $\lambda$ for which the proxy approves such (noise-augmented) points rarely enough, at the corrected level $\lowerval(\alpha,\gB^{-1})$. At test time only the proxy is evaluated.
The calibration produces a scalar $\lambda^*$ for the given radius $r$. At test time, for the input $\xtestpert$, one noise sample from $\psi$ is drawn, added to $\xtestpert$, and the result is passed through the proxy. Based on the desired guarantee our inference can be in either of the following directions:
\begin{enumerate*}
    \item[1)] If we target \autoref{eq:proxy-risk-solid}, calibration uses a sound certificate as backbone to identify non-robust points; at deployment the same inference rule produces the label $y$ and the proxy certifies robustness -- therefore, we follow the inference rule of the certificate (in our experiments a majority vote over 50 noisy samples);
    \item[2)] If we target \autoref{eq:proxy-risk-augmented} a single noise augmentation from $\psi$ is used both to obtain the predicted label and to evaluate the proxy.
\end{enumerate*}
In all cases, the output is a (label, certification decision) pair. Pseudo-code for all three variants is in \autoref{sec:algorithm}. Based on the budget for calibration, \method-A (\autoref{thrm:proxy-cert-augmented}) is the cheapest option with single sample both in calibration and test, \method (\autoref{thrm:proxy-cert}) stays in between only sampling to compute the backbone certificate, and \method-H uses high sample rate both for the backbone certificate and for estimation of the worst case risk. All three methods use a single sample at test time; see \autoref{tab:guarantees} in \autoref{sec:algorithm}.

\paragraph{Fundamental limitations.}
Our guarantee relaxes the soundness to bounded error -- our proxy certifies by mistake with a rate of at most $\alpha$ for either clean or perturbed inputs. So to apply our framework we assume that such bounded error might be tolerable; e.g. large-scale dataset auditing, robustness monitoring, high-throughput content pipelines, or APIs where the proxy decides whether to escalate to the expensive certificate. Additionally, the backbone certified ratio is a performance ceiling;
the \emph{sound} certified ratio of \method (the fraction of certifications that are correct) is bounded by the backbone certificate's certified ratio; however, the total certified ratio (because of the at-most-$\alpha$ fraction of permitted false approvals) can get slightly higher (see \autoref{tab:imagenet-main} ViT at $r{=}0$).
Our framework relies on a holdout calibration set $\dcal=\{\vx_i\}_{i=1}^n$ sampled exchangeably from the distribution of clean future test inputs; in contrast to conformal prediction, this assumption is relatively mild because the calibration points are unlabeled. Nevertheless, our approach is similarly sensitive to shifts in the distribution of unobserved clean points. Under distribution shift, the proxy can be recalibrated on a few recent unlabeled inputs by running the expensive certificate only on them. \method offers a marginal guarantee rather than conditional. This mirrors the usual limitation of conformal prediction, and risk control. This means that the proxy may be more conservative on some parts of the distribution and less reliable on others. Our main assumption is that the clean inputs follow the deployment distribution. This means that the adversary is not in control of data distribution (which is a shift in subpopulation). In local regions, and in low chance classes the risk can exceed $\alpha$  (we report class-specific risks in \autoref{tab:stratified}, and \autoref{tab:stratified-cifar}).

These limitations motivate the future work: \begin{enumerate*}[label=(\roman*),itemjoin={{; }},itemjoin*={{; and }}]
    \item designing better proxies to increase coverage over the backbone's certified ratio
    \item defining localized and adaptive forms of CRC, approaching for close-to-conditional guarantees
    \item handling distribution shift in the latent clean space.
\end{enumerate*}

\section{Experiments}
\label{sec:experiments}

\begin{figure}
    \centering
    \input{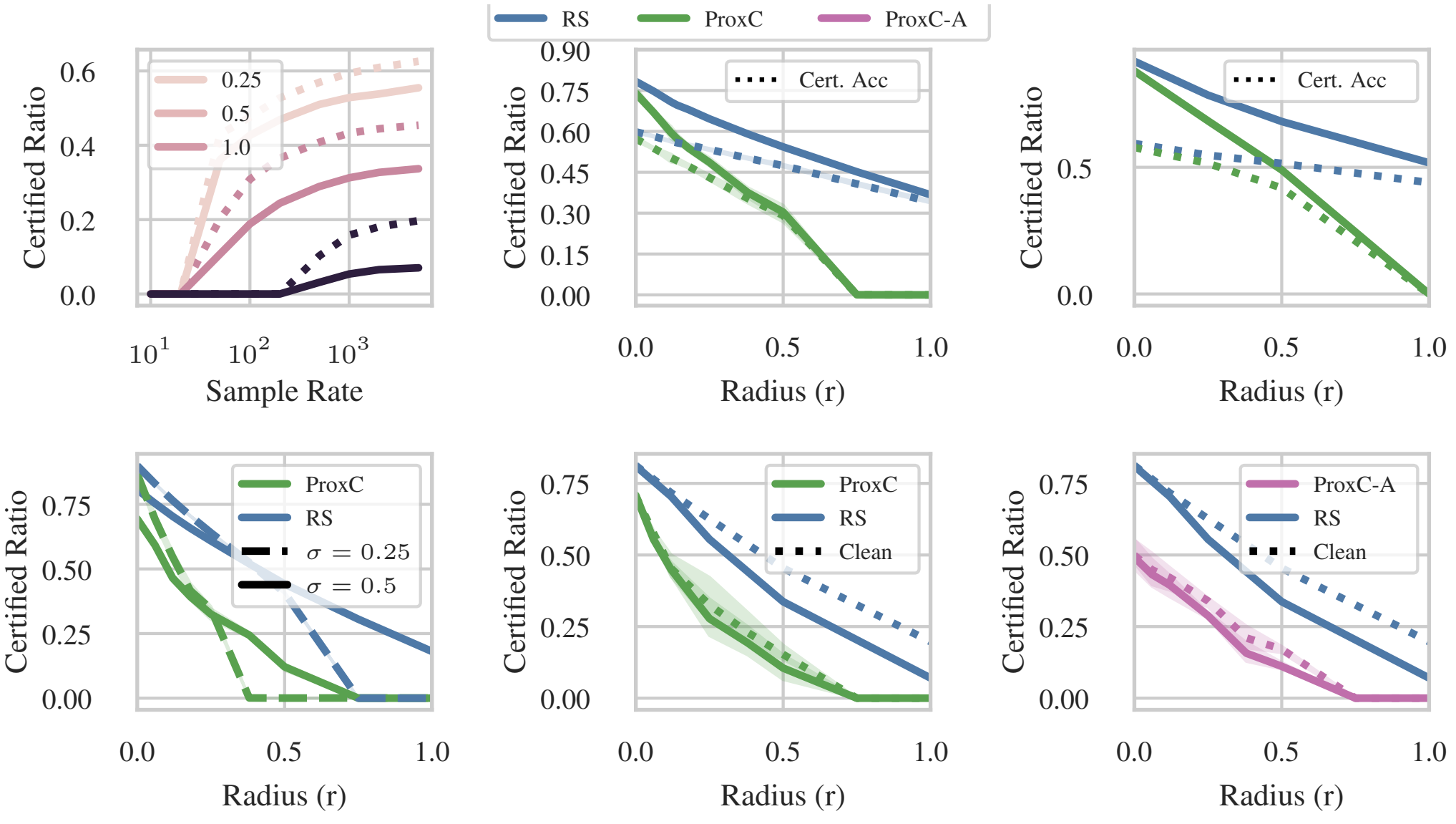}
    \caption{\emph{Top row (ImageNet).} Left: certified ratio of randomized smoothing (RS) as a function of the Monte Carlo sample budget, across perturbation radii $r$ (dotted lines show the performance on perturbed points). Middle: certified ratio (solid) and certified accuracy (dotted) as functions of $r$ for ResNet. Right: the same comparison on the Vision Transformer. \emph{Bottom row (CIFAR-10).} Left: Effect of $\sigma$ on certified ratio. Middle: certified ratio of \method, and Right \method-A across various radii. All results are with $\sigma=0.5$ unless specified. All curves report mean over 5 runs.}
    \label{fig:imagenet-general}
\end{figure}



Our evaluation focuses on three points, also summarized in \autoref{tab:timing}. We show that \begin{enumerate*}[label=(\roman*),itemjoin={{; }},itemjoin*={{; and }}]
\item as guaranteed, the marginal false-approval rate remains below the tolerance $\alpha$ on both clean and adversarial inputs
\item \method recovers a large fraction of the certified ratio and accuracy of randomized smoothing at a substantially lower computational cost
\item the computational gains scale with the reduction in Monte Carlo samples required at inference time
\end{enumerate*} We further compare \method with state-of-the-art robustness certificates beyond randomized smoothing. Despite relaxing soundness to bounded risk, randomized smoothing scales more effectively to larger models such as vision transformers. We evaluate both text and image settings. The code is available at \url{https://github.com/soroushzargar/ProxC}.


\paragraph{Expected behaviour.}
The objective of \method is not to outperform the backbone certificate on which it is calibrated; by design, the backbone is the ceiling for performance. Instead, our aim is to approximate the certificate with a cheaper function (without the need for MC-sampling), while controlling its false signals statistically. In this sense, \method is successful when it remains valid, and can cover a higher portion of backbone certificate's approval ratio; by design it is also supposed to reduce the cost by thousands of forward passes. Therefore we compare the risk as a sanity check, and report the certified ratio and accuracy as the analog to coverage in conformal prediction.

\paragraph{Setup of image experiments. }
We evaluate \method on CIFAR-10 \citep{krizhevsky2009learning} and ImageNet \citep{deng2009imagenet} datasets under $\ell_2$ perturbations. Our results are applicable to any binary certificate, and any smoothing scheme (same as \citet{Zargarbashi2025OneSI}). We use two main model configurations. 
\begin{enumerate*}[label=(\roman*)]
    \item ResNet-110 (CIFAR-10) and ResNet-50 (ImageNet) from \citet{cohen2019certified}, trained with Gaussian noise augmentation at the same $\sigma$ as smoothing (default is $\sigma=0.5$).
    \item Following \citet{salman2020denoised}, and \citet{carlini2022certified}, the augmented input ($\xtestpert + \vepsilon$) is passed through class-unconditional diffusion followed by a BEiT-L model from \citet{bao2021beit}. 
    Models are frozen; only the proxy threshold $\lambda$ is calibrated.
\end{enumerate*}
Our default tolerance level is $\alpha=0.05$ and for calibration-heavy experiment we use the Clopper-Pearson intervals with union confidence of $1-\delta=0.999$ --- both for randomized smoothing and our own proxy certificate.\footnote{We assume the certificates are $100\%$ correct, meaning that while computing the risk, at most $\delta = 0.1\%$ of the calibration points mistakenly report zero as error. Replacing with the upper bound risk for them, the overall risk will increase to at most $\le \alpha + \frac{\delta\cdot n}{n+1} \le \alpha + \delta$ which here is $5.1\%$.} We report the average over 5 runs with calibration set resampling (unless specified). $|\dcal|$ is set variably between 200 and 500 points. We examine our method on 4096 images from CIFAR-10 and 2048 images from ImageNet. By default we run the backbone certificate with 2000 MC samples (unless specified), out of which 50 are used for the majority-vote prediction and 1950 for the certificate. Note that \method-A does not require such sampling. For calibration heavy methods, we use $20{,}000$ MC samples at calibration-time which is a one time pre-processing.

\paragraph{Cost accounting.}
The single forward pass refers only to the certification decision: \method and \method-H still obtain the label with the inference rule of the backbone ($50$ forwards, i.e. $51$ in total), \method-A uses the same noisy forward for the label and the approval ($1$ in total). The full accounting, the fallback cost across radii, and the break-even point of \method-H are in \autoref{tab:cost}, \autoref{tab:fallback}, and \autoref{sec:expr-appendix}.

\begin{table}[t]
\centering
\caption{Certified ratio ({CR}, \%) and certified accuracy ({CA}, \%) on
  ImageNet with  $\sigma=0.5$.
  RS uses  $n=2{,}000$  (ResNet-50) and  $n=10{,}000$  (ViT) Monte Carlo samples
  (single calibration run each). 
  Entries marked  $\dagger$  use deterministic (non-smoothing) certificates reported directly from the reference. Note the comparison is between pointwise and marginal guarantees.}
\label{tab:imagenet-main}
\vspace{-0.5em}
\scriptsize
\setlength{\tabcolsep}{7.0pt}
\renewcommand{\arraystretch}{0.82}
\begin{tabular}{l c rr rrr r}
\toprule
&& \multicolumn{2}{c}{\textbf{RS} (Pointwise)} &
\multicolumn{3}{c}{\textbf{ProxC} (Marginal 5\% Risk)} &
\textbf{Ref.\ CA} \\
\cmidrule(lr){3-4}\cmidrule(lr){5-7}
\textbf{Model} & $r$ & CR & CA & CR & CA & Risk & $(\dagger)$ \\
\midrule
\multirow{5}{*}{ResNet-50}
& \num{0.00} & \num{78.79} & \num{60.00} & \num{72.88} & \num{56.78} & \num{4.17} & --- \\
& \num{0.14} & \num{70.06} & \num{56.08} & \num{56.11} & \num{48.53} & \num{2.29} & --- \\
& \num{0.25} & \num{64.91} & \num{53.61} & \num{46.29} & \num{41.70} & \num{1.25} & --- \\
& \num{0.50} & \num{54.65} & \num{47.62} & \num{29.01} & \num{28.04} & \num{0.26} & --- \\
& \num{0.75} & \num{45.54} & \num{41.12} & \multicolumn{3}{c}{---} & --- \\
\midrule
\multirow{4}{*}{ViT}
& \num{0.00} & \num{91.81} & \num{59.42} & \num{95.00} & \num{59.62} & \num{5.79} & --- \\
& \num{0.25} & \num{78.96} & \num{55.23} & \num{58.56} & \num{46.96} & \num{1.83} & --- \\
& \num{0.50} & \num{68.44} & \num{51.60} & \num{37.54} & \num{34.04} & \num{0.31} & --- \\
& \num{0.75} & \num{59.77} & \num{48.19} & \multicolumn{3}{c}{---} & --- \\
\midrule
\multicolumn{8}{l}{\textit{Verifier-based certificates $(\dagger)$: $\epsilon=36/255\approx0.14$, CA only, pointwise guarantee}} \\
\midrule
LipNeXt-2B$^\dagger$       & \num{0.14} & \multicolumn{5}{c}{--- \cite{lipnext2026} ---} & \num{41.2} \\
LipNeXt-1B$^\dagger$       & \num{0.14} & \multicolumn{5}{c}{--- \cite{lipnext2026} ---} & \num{40.3} \\
LiResNet+EMMA$^\dagger$    & \num{0.14} & \multicolumn{5}{c}{--- \cite{hu2023unlocking} ---} & \num{35.0} \\
CertViT (ViT-T/16)$^\dagger$ & \num{0.14} & \multicolumn{5}{c}{--- \cite{certvit2023} ---} & \num{21.7} \\
CertViT (DeiT-T)$^\dagger$ & \num{0.14} & \multicolumn{5}{c}{--- \cite{certvit2023} ---} & \num{17.3} \\
\bottomrule
\end{tabular}
\vspace{-0.7em}
\end{table}


\begin{table}[t]
\centering
\caption{%
  Calibration sample budget $n_\mathrm{mc}$ for {ProxC-H} on ImageNet (ViT + denoising, $\sigma\!=\!0.5$, $\alpha\!=\!0.05$,
  averaged over 3 runs).
  Each cell reports CR\,(\%) with FA\,(\%) in
  parentheses.
  All FA values remain $\leq\alpha=5\%$.%
}
\label{tab:ablation-calheavy}
\setlength{\tabcolsep}{5pt}
\small
\begin{tabular}{c cc ccccc}
\toprule
 & \multicolumn{2}{c}{Base (Cohen)} &
   \multicolumn{5}{c}{\textbf{ProxC-H} Proxy CR (FA)} \\
\cmidrule(lr){2-3}\cmidrule(lr){4-8}
$r$ & CR & CA & $K\!=\!1\mathrm{k}$ & $K\!=\!2.5\mathrm{k}$ & $K\!=\!5\mathrm{k}$ & $K\!=\!10\mathrm{k}$ & $K\!=\!20\mathrm{k}$ \\
\midrule
\num{0.25} & \num{79.0} & \num{55.2} & \num{62.2}\,(\num{2.6}) & \num{63.9}\,(\num{2.9}) & \num{64.6}\,(\num{3.0}) & \num{63.2}\,(\num{2.4}) & \num{63.6}\,(\num{2.6}) \\
\num{0.38} & \num{73.4} & \num{53.3} & \num{51.8}\,(\num{1.5}) & \num{54.4}\,(\num{2.0}) & \num{55.7}\,(\num{2.3}) & \num{55.5}\,(\num{1.5}) & \num{56.0}\,(\num{1.6}) \\
\num{0.50} & \num{68.4} & \num{51.6} & \num{39.2}\,(\num{0.4}) & \num{46.1}\,(\num{0.9}) & \num{48.1}\,(\num{1.1}) & \num{49.1}\,(\num{1.0}) & \num{49.8}\,(\num{1.0}) \\
\num{0.75} & \num{59.8} & \num{48.2} & --- & --- & \num{21.5}\,(\num{0.0}) & \num{35.2}\,(\num{0.2}) & \num{38.9}\,(\num{0.5})\\
\bottomrule
\end{tabular}

\end{table}

\paragraph{Results on image datasets.}
\autoref{tab:timing} reports results on ImageNet at $r\!=\!0.14$. There are two use cases: using \method as standalone reduces the certification cost from the MC sampling budget to 1, recovering acceptably high portion of the certified accuracy (in our case it is a $1950\!\times$ reduction of the certification-only cost, and from 2000 to 51 forwards, i.e. $39\times$, end-to-end including the label). In fallback setup, the full certified ratio of the backbone certificate is recovered, at the cost of the speed up. Still in this setup we have a speed up of $2.50\times$ in certification-only cost ($2.41\times$ end-to-end), which decreases with $r$ (\autoref{tab:fallback}).
\method-H improves the certified ratio of the ViT pipeline at $r\!=\!0.5$ at the cost of a heavier calibration phase (\autoref{tab:imagenet_sigma05}). It uses the heavy computation only once, and in the test time it is under the same single-sample regime.
The empirical false-approval rate remains well below $\alpha\!=\!5\%$ in almost all experiments, confirming the validity of the theoretical guarantee  (the guarantee is in expectation also over calibration sets, meaning that for a single split exceeding $\alpha$ by a minor margin can happen, e.g. ViT at $r\!=\!0$ in \autoref{tab:imagenet-main}; this variance shrinks as $|\dcal|$ grows, see \autoref{tab:calsize}).
\autoref{fig:imagenet-general} compares \method, and \method-A with the backbone randomized smoothing certificate in more detail: It shows that with RS we can not reduce the sample rate to acceptable budgets (top left). It also shows the effect of the smoothing magnitude $\sigma$ (bottom left).

Our main goal can be empirically validated in \autoref{tab:imagenet-main}. Randomized smoothing, and exact certificates operate at two ends of a trade-off between computational efficiency (from sample-rate), and other benefits like larger robust radii, and black-box access. The certification step of \method operates at the same computational cost as verification (a single forward), while its guarantee is from randomized smoothing. Setting aside the relaxation from a sound certificate to marginal bounded error, it highlights the potentials of randomized smoothing once its sampling bottleneck is removed by the proxy --- redefining the trade-off. This relaxation emphasizes that this comparison should be interpreted with care -- the results are not directly comparable; but it shows the trade-off between soundness, cost, and ratio. In applications where this relaxation is acceptable, the savings can be decisive. As noted before, the Jensen's inequality is the cost of using a single sample in calibration. \method-H instead uses sampling to estimate the expected risk. As shown in \autoref{tab:ablation-calheavy}, by increasing the calibration time sample rate even new values of $r$ become reachable. Note that this sampling is only done once, during calibration. Later in \autoref{tab:cifar-attacked} we show the validity under PGD-RS attack which aims to break randomized smoothing. Still under attack the risk remains below the tolerance level.

\paragraph{NLP experiments.}
We validate the framework on text classification, using CERT-ED \citep{huang2024cert}.
The corresponding threat model is the edit (Levenshtein) distance at the token level 
The backbone smoothed classifier is RoBERTa-base fine-tuned on AG-News (4-class topic classification, 7{,}600 test inputs) with the matching noise distribution applied at every training forward pass.
\autoref{tab:proxy-results} reports the performance of \method, and \method-H alongside the base certificate. Here, the backbone certificate is over edit distance, and the proxy certificate is over $r$ insertions.
We further discuss the theoretical background behind the NLP-experiments in \autoref{sec:expr-appendix}.

\begin{table}[t]
\centering
\caption{NLP task: Certified ratio (CR) and certified accuracy (CA) of the backbone certificate and \method and risk. The certification is over $r$-insertions.}
\label{tab:proxy-results}
\setlength{\tabcolsep}{6pt}
\renewcommand{\arraystretch}{1.08}
\small
\begin{tabular}{llrrrrrr}
\toprule
 &  & \multicolumn{6}{c}{$r$} \\
\cmidrule(lr){3-8}
Method & Metric & \num{0} & \num{1} & \num{2} & \num{3} & \num{4} & \num{5} \\
\midrule
\multirow{2}{*}{Backbone}
  & CR & \num{98.5}\% & \num{89.3}\% & \num{77.9}\% & \num{61.8}\% & \num{38.8}\% & \num{14.6}\% \\
  & CA & \num{91.0}\% & \num{84.7}\% & \num{75.2}\% & \num{60.3}\% & \num{38.2}\% & \num{14.5}\% \\
\midrule
\multirow{2}{*}{ProxC}
  & CR & \num{100.0}\% & \num{70.7}\% & \num{45.2}\% & \num{23.9}\% & \num{10.3}\% & \num{4.5}\% \\
  & CA & \num{92.0}\%  & \num{66.3}\% & \num{43.6}\% & \num{23.5}\% & \num{10.3}\% & \num{4.5}\% \\
\midrule
Risk
  & -- & \num{1.5}\% & \num{4.5}\% & \num{4.1}\% & \num{3.7}\% & \num{3.7}\% & \num{3.3}\% \\
\bottomrule
\end{tabular}
\end{table}

\section{Conclusion}
We introduced the proxy certificate (\method), a framework that significantly saves computational cost of robustness, by trading the pointwise soundness of a robustness certificate for a bounded marginal false-approval rate. We combine randomized smoothing with conformal risk control, allowing for a robust risk control setup, that works with a single augmented sample for the certification decision at test time (in contrast with heavy MC sampling). Depending on the compute budget, the calibration can be done with heavy MC sampling or again with a single sample. As a standalone certificate, \method certifies with one forward pass (51 end-to-end including the label), while as a fallback it recovers the backbone's certified ratio with a more modest speedup ($1.3$--$3.8\times$ end-to-end on ImageNet). Under the bounded-risk relaxation, \method
achieves higher certified accuracy than state-of-the-art deterministic verifiers at matching certification
cost, although the two provide qualitatively different guarantees (marginal vs.\ pointwise) on different architectures, while its black-box nature enables application to large architectures such as vision transformers
and language models that are out of reach for white-box verification. This highlights the potential in randomized smoothing, as a powerful defense. Our guarantee is marginal and does not extend to individual inputs, classes, or shifted distributions; addressing these, together with a cheaper label prediction and early-stopping backbones, is left for future work.

\begin{ack}
This work was supported only by the authors' affiliations, the CISPA Helmholtz Center for Information Security and the University of Cologne, and received no third-party funding. The authors declare no competing interests.
\end{ack}

\bibliographystyle{plainnat}
\bibliography{main}


\newpage
\appendix

\section{Algorithms}
\label{sec:algorithm}

For simplicity, in the algorithms below we write $\hat h(\vz,\lambda;y)$ for the proxy approval of class $y$ at the received point $\vz$. The class $y$ can be determined by any arbitrary inference rule. For \method, and \method-H, we use the conventional classification approach in randomized smoothing, which is to take the majority vote class over a low sample rate (e.g. 50). We also write
\[
    \bar R_n(\lambda)=\frac{1+\sum_{i=1}^n R_i(\lambda)}{n+1},
\]
which is the conformal empirical risk with the standard one-point correction. For \method, and \method-H we use $\lowerval(\alpha,\gB^{-1})$, and $\alpha-\delta$ as the risk threshold, respectively, following the setup in \autoref{thrm:robust-rs-risk}.

\autoref{alg:proxc} is the pseudo-code for \method. It uses the backbone certificate only on $\dcal$ in order to identify the points for which false approval is possible -- the points that the backbone certificate approves directly have a risk of 0 (i.e. $c_i=1$) since for any value of the proxy they are not a false approval. On non-certified points, the proxy is allowed to approve only often enough that the smoothed, robustly corrected CRC risk stays below $\alpha$. At test time the certificate is no longer evaluated: the calibrated proxy itself is the certificate.

\begin{algorithm}[h]
\caption{\textsc{ProxC}: Proxy Certificate Inference}
\label{alg:proxc}
\begin{algorithmic}[1]
\Require Calibration set $\dcal=\{\vx_i\}_{i=1}^n$, classifier $f$, sound certificate $\cert{f,r}$, smoothing scheme $\psi$, proxy $\hat h$, target risk $\alpha$, radius $r$
\Ensure Calibrated threshold $\lambda^*$ and proxy certificate $\certproxy_{f,r}$
\State \textbf{Calibration:}
\For{$i=1,\ldots,n$}
    \State Compute the reference prediction $y_i$ using the inference rule associated with $\cert{f,r}$
    \State Compute the certificate decision $c_i \gets \cert{f,r}(\vx_i)$
    \State Draw one independent smoothing noise $\vepsilon_i\sim\psi$ and set $\vz_i\gets \vx_i+\vepsilon_i$
\EndFor
\State Define $R_i(\lambda)\gets (1-c_i)\hat h(\vz_i,\lambda;y_i)$
\State Via binary search, set
\[
    \lambda^*
    \gets
    \inf\left\{\lambda:
    \bar R_n(\lambda)\le \lowerval(\alpha,\gB^{-1})
    \right\}.
\]
\State \textbf{Deployment:}
\State Given $\tilde\vx$, compute $y\gets$ the reference prediction at $\tilde\vx$ (e.g.\ majority vote over 50 noisy samples)
\State Draw $\vepsilon\sim\psi$ and return $y$ if $\hat h(\tilde\vx+\vepsilon,\lambda^*;y)=1$; otherwise abstain
\State \Return $\certproxy_{f,r}(\tilde\vx)=\hat h(\tilde\vx+\vepsilon,\lambda^*;y)$
\end{algorithmic}
\end{algorithm}

\begin{algorithm}[t]
\caption{\textsc{ProxC-A}: Proxy-Augmented Inference}
\label{alg:proxca}
\begin{algorithmic}[1]
\Require Calibration set $\dcal=\{\vx_i\}_{i=1}^n$, classifier $f$, smoothing scheme $\psi$, proxy $\hat h$, target risk $\alpha$, radius $r$
\Ensure Calibrated threshold $\lambda^*$ and augmented proxy certificate $\certproxy_{f,r}^{\mathrm{A}}$
\State \textbf{Calibration:}
\For{$i=1,\ldots,n$}
    \State Compute the clean anchor prediction $y_i\gets f(\vx_i)$
    \State Draw one independent smoothing noise $\vepsilon_i\sim\psi$ and set $\vz_i\gets \vx_i+\vepsilon_i$
    \State Compute the augmented prediction $\tilde y_i\gets f(\vz_i)$
\EndFor
\State Define $R_i(\lambda)\gets \sI\{\tilde y_i\neq y_i\}\hat h(\vz_i,\lambda;\tilde y_i)$
\State Set
\[
    \lambda^*
    \gets
    \inf\left\{\lambda:
    \bar R_n(\lambda)\le \lowerval(\alpha,\gB^{-1})
    \right\}.
\]
\State \textbf{Deployment:}
\State Given $\tilde\vx$, draw $\vepsilon\sim\psi$, set $\vz\gets \tilde\vx+\vepsilon$, and compute $y\gets f(\vz)$
\State Return $y$ if $\hat h(\vz,\lambda^*;y)=1$; otherwise abstain
\State \Return $\certproxy_{f,r}^{\mathrm{A}}(\tilde\vx)=\hat h(\tilde\vx+\vepsilon,\lambda^*;f(\tilde\vx+\vepsilon))$
\end{algorithmic}
\end{algorithm}

\autoref{alg:proxca} removes the explicit certificate completely. Instead of asking whether $\cert{f,r}$ can prove local stability, calibration directly measures whether the randomized inference rule changes the clean prediction. The proxy is then calibrated as an abstention rule for the augmented probabilistic classifier itself. This is the cheapest version of the framework: calibration and deployment both use the same one-sample augmented inference mechanism, and the guarantee is exactly the joint event of toggled prediction and false-approval in \autoref{eq:proxy-risk-augmented}.

\begin{algorithm}[t]
\caption{\textsc{ProxC-H}: Calibration-Heavy Proxy Certificate}
\label{alg:proxch}
\begin{algorithmic}[1]
\Require Calibration set $\dcal=\{\vx_i\}_{i=1}^n$, classifier $f$, sound certificate $\cert{f,r}$, smoothing scheme $\psi$, proxy $\hat h$, target risk $\alpha$, radius $r$, tuning sample-rate $m_\mathrm{tune}$, certification sample-rate $m_\mathrm{cert}$, confidence parameter $\delta$, tuning slack $\delta_0>0$
\Ensure Calibrated threshold $\lambda^*$ and proxy certificate $\certproxy_{f,r}^{\mathrm{H}}$
\State \textbf{Precomputation:}
\For{$i=1,\ldots,n$}
    \State Compute the reference prediction $y_i$ and certificate decision $c_i\gets\cert{f,r}(\vx_i)$
\EndFor
    \State $\lambda_\mathrm{cand}\gets\Call{Tune}{\dcal,\,y_{1:n},\,c_{1:n},\,m_\mathrm{tune},\,m_\mathrm{cert},\,\alpha,\,\delta,\,\delta_0}$
    \If{$\Call{Validate}{\dcal,\,y_{1:n},\,c_{1:n},\,\lambda_\mathrm{cand},\,m_\mathrm{cert},\,\alpha,\,\delta}$}
    \State $\lambda^*\gets\lambda_\mathrm{cand}$
    \Else
        \State $\lambda^*\gets1$ \Comment{$\hat h(\cdot,1)=0$: no certifications}
    \EndIf
    \State \textbf{Deployment:}
    \State Given $\tilde\vx$, compute $y\gets$ the reference prediction at $\tilde\vx$ (e.g.\ vote over 50 noisy samples)
    \State Draw one smoothing noise $\vepsilon\sim\psi$ and return $y$ if $\hat h(\tilde\vx+\vepsilon,\lambda^*;y)=1$; otherwise abstain
    \State \Return $\certproxy_{f,r}^{\mathrm{H}}(\tilde\vx)=\hat h(\tilde\vx+\vepsilon,\lambda^*;y)$
\end{algorithmic}
\end{algorithm}

\begin{table}[h]
\centering
\caption{What is guaranteed by randomized smoothing (RS) and by the \method variants. In all cases the adversary knows the model weights, the clean input, and the procedure, and may choose any $\xtestpert\in\gB(\xtest)$; it does not control or observe the internal randomness (e.g.\ $\vepsilon$).}
\label{tab:guarantees}
\footnotesize
\setlength{\tabcolsep}{3pt}
\renewcommand{\arraystretch}{1.0}
\begin{tabular}{p{1.1cm}p{2.6cm}p{2.6cm}p{2.5cm}p{3.7cm}}
\toprule
Method & Prediction rule & Randomness & Averaged over & Guarantee \\
\midrule
RS & Majority vote, small $n_0$ & MC-corrected & --- (pointwise) & Pointwise soundness \\
\addlinespace[2pt]
\method & Same as RS & One fresh smoothing noise at proxy evaluation & Clean input, calibration set, internal noise & Marginal false-approval control (not pointwise) \\
\addlinespace[2pt]
\method-H & Same as \method & One fresh noise draw at test & Same as \method, plus calibration MC & Same marginal guarantee, usually less conservative \\
\addlinespace[2pt]
\method-A & One-sample noise-augmented inference; approves with the same draw & One fresh draw affects both prediction and approval & Clean input, calibration set, noises & Controls the joint event \{randomized prediction toggles\} $\wedge$ \{proxy approves\}; not a certificate for a fixed deterministic prediction \\
\bottomrule
\end{tabular}
\end{table}

\subsection{Front-Loaded Calibration for \method-H}
\label{sec:flor-proxch}

\autoref{alg:proxch} is useful when a single-time high computational cost is available; given higher compute power, it allows for less conservative proxy. Calibration proceeds in two independent phases (detailed below): a tuning phase (\autoref{alg:proxch-tune}) that uses $m_\mathrm{tune}$ noise samples per point to select a candidate threshold $\lambda_\mathrm{cand}$ via a simulated CRC condition, and a validation phase (\autoref{alg:proxch-validate}) that uses $m_\mathrm{cert}$ fresh, independent samples to confirm $\lambda_\mathrm{cand}$ at the fixed threshold. The two-stage split eliminates the selection effect that arises when the same samples are used to both scan for $\lambda^*$ and bound the risk at $\lambda^*$. Deployment remains identical to \textsc{ProxC}: one noisy proxy evaluation decides whether the prediction is accepted.

\begin{algorithm}[t]
\caption{\textsc{ProxC-H-Tune}: Tuning Phase for \textsc{ProxC-H}}
\label{alg:proxch-tune}
\begin{algorithmic}[1]
\Require {Precomputed $(y_i,c_i)_{i=1}^n$, calibration set $\dcal$, proxy $\hat h$, smoothing $\psi$, $m_\mathrm{tune}$, $m_\mathrm{cert}$, $\alpha$, $\delta$, tuning slack $\delta_0$}
\Ensure {Candidate threshold $\lambda_\mathrm{cand}$}
\For{{$i=1,\ldots,n$}}
    \For{{$j=1,\ldots,m_\mathrm{tune}$}}
        \State {Draw $\vepsilon_{i,j}\sim\psi$;\; set $a_{i,j}(\lambda)\gets\hat h(\vx_i+\vepsilon_{i,j},\lambda;\,y_i)$}
    \EndFor
    \State {$\hat p_i^{\mathrm{tune}}(\lambda)\gets m_\mathrm{tune}^{-1}\sum_{j=1}^{m_\mathrm{tune}} a_{i,j}(\lambda)$}
    \State {$k_i^{\mathrm{sim}}(\lambda)\gets \lfloor\hat p_i^{\mathrm{tune}}(\lambda)\cdot m_\mathrm{cert}\rfloor$}
    \State {$\overline p_i^{\mathrm{sim}}(\lambda)\gets\mathrm{CP}^{\uparrow}\!\bigl(k_i^{\mathrm{sim}}(\lambda),\,m_\mathrm{cert},\,1-\delta/n\bigr)$}
    \State {$R_i^{\mathrm{sim}}(\lambda)\gets (1-c_i)\,\upperval\!\bigl(\overline p_i^{\mathrm{sim}}(\lambda),\,\gB(\vx_i)\bigr)$}
\EndFor
\State {$\bar R_n^{\mathrm{sim}}(\lambda)\gets\frac{1+\sum_{i=1}^n R_i^{\mathrm{sim}}(\lambda)}{n+1}$}
\State {\Return $\lambda_\mathrm{cand}\gets\inf\!\left\{\lambda:\bar R_n^{\mathrm{sim}}(\lambda)\le\alpha-\delta-\delta_0\right\}$}

\end{algorithmic}
\end{algorithm}

\begin{algorithm}[t]
\caption{{\textsc{ProxC-H-Validate}: Validation Phase for \textsc{ProxC-H}}}
\label{alg:proxch-validate}
\begin{algorithmic}[1]
\Require {Precomputed $(y_i,c_i)_{i=1}^n$, calibration set $\dcal$, proxy $\hat h$, smoothing $\psi$, fixed threshold $\lambda_\mathrm{cand}$, $m_\mathrm{cert}$, $\alpha$, $\delta$}
\Ensure {Boolean: $\mathsf{True}$ iff the CRC condition holds at $\lambda_\mathrm{cand}$}
\For{{$i=1,\ldots,n$}}
    \For{{$j=1,\ldots,m_\mathrm{cert}$}}
        \State {Draw $\vepsilon_{i,j}^{\mathrm{cert}}\sim\psi$;\; set $a_{i,j}^{\mathrm{cert}}\gets\hat h(\vx_i+\vepsilon_{i,j}^{\mathrm{cert}},\lambda_\mathrm{cand};\,y_i)$}
    \EndFor
    \State {$k_i^{\mathrm{cert}}\gets\sum_{j=1}^{m_\mathrm{cert}} a_{i,j}^{\mathrm{cert}}$}
    \State {$\overline p_i^{\mathrm{cert}}\gets\mathrm{CP}^{\uparrow}\!\bigl(k_i^{\mathrm{cert}},\,m_\mathrm{cert},\,1-\delta/n\bigr)$}
    \State {$R_i^{\mathrm{cert}}\gets (1-c_i)\,\upperval\!\bigl(\overline p_i^{\mathrm{cert}},\,\gB(\vx_i)\bigr)$}
\EndFor
\State {$\bar R_n^{\mathrm{cert}}(\lambda_\mathrm{cand})\gets\frac{1+\sum_{i=1}^n R_i^{\mathrm{cert}}}{n+1}$}
\State {\Return $\bigl[\bar R_n^{\mathrm{cert}}(\lambda_\mathrm{cand})\le \alpha-\delta\bigr]$}
\end{algorithmic}
\end{algorithm}

Taken together, the three algorithms give a simple recipe. Use \textsc{ProxC} when a trusted certificate can be afforded once on the calibration set and deployment must be cheap. Use \textsc{ProxC-A} when even calibration-time certification is undesirable and one is willing to certify the randomized inference rule directly. Use \textsc{ProxC-H} when calibration is allowed to be sample-heavy; the additional calibration samples usually reduce the conservativeness of the single-sample proxy while preserving the same marginal false-approval guarantee.

\paragraph{The multiple-testing issue in single-phase calibration.}
The probabilities used in \autoref{eq:proxy-crc-lambda-heavy} need to be estimated via MC sampling. However, using an algorithm like binary search over the same set of samples brings statistical flaw to the estimation. On the other hand, resampling and re-estimating the values at each step of the binary search is also expensive. As an alternative, we divide the computation of $\lambda^*$ into two phases: tuning (with low sample rate), and validation with high sample rate. The theoretical guarantee is the result of the validation step, therefore we can reuse the samples while tuning. 

\paragraph{Two-stage fix.}
We decouple threshold selection from threshold validation by splitting the noise budget as $m_\mathrm{tune} + m_\mathrm{cert}$. The \emph{tuning phase} (\autoref{alg:proxch-tune}) uses $m_\mathrm{tune}$ samples per calibration point to select a candidate $\lambda_\mathrm{cand}$ by simulating the certification budget: the empirical count $\hat p_i^{\mathrm{tune}}(\lambda)\cdot m_\mathrm{cert}$ is projected to scale $m_\mathrm{cert}$, a Clopper-Pearson bound is computed at that simulated count, and the smoothing upper bound is applied, all with an extra slack $\delta_0$. The \emph{validation phase} (\autoref{alg:proxch-validate}) then draws $m_\mathrm{cert}$ independent noise samples at the \emph{fixed} threshold $\lambda_\mathrm{cand}$. Because $\lambda_\mathrm{cand}$ is determined before the validation samples are drawn, the CP bounds in the validation phase are valid for the specific threshold $\lambda_\mathrm{cand}$ without any selection correction. Note that both phases reuse the same calibration inputs $\vx_1,\dots,\vx_n$. This is not a source of selection bias; only reusing the same Monte Carlo noise to both select and validate $\lambda_\mathrm{cand}$ would be. Splitting the calibration points into tuning and validation sets is therefore unnecessary, and would only reduce the effective calibration size.

\paragraph{Formal validity.}
\begin{proposition}
\label{prop:flor-validity}
Let $\lambda^*$ be the output of \autoref{alg:proxch}, i.e. $\lambda^*=\lambda_\mathrm{cand}$ if the validation check in \autoref{alg:proxch-validate} passes and $\lambda^*=1$ otherwise, and suppose the validation noise is drawn independently of the tuning noise. Then
\[
    \mathbb{E}\!\left[(1-c(\xtest))\,\upperval\!\bigl(p(\xtest,\lambda^*),\,\gB\bigr)\right] \le \alpha,
\]
where $p(\vx,\lambda)=\Pr_{\vepsilon}[\hat h(\vx+\vepsilon,\lambda)=1]$, $c(\vx)=\cert{f,r}(\vx)$, and the expectation is over the calibration set, the test point, and the tuning and validation noise.
\end{proposition}
\begin{proof}
Let $R^\circ(\vx,\lambda)=(1-c(\vx))\,\upperval(p(\vx,\lambda),\gB)$ be the exact (infinite-MC) risk. It is bounded in $[0,1]$, non-increasing in $\lambda$ (since $p$ is non-increasing in $\lambda$ and $\upperval$ is non-decreasing in its first argument), and $R^\circ(\cdot,1)=0$. The backbone decisions $c_i$ (computed with their own independent MC noise per point) are treated as part of the exchangeable calibration data.

\emph{Oracle step.} Consider first an oracle procedure that knows $p$ exactly and targets $\alpha-\delta$:
\[
    \lambda_\mathrm{or}=\inf\left\{\lambda:\frac{1+\sum_{i=1}^n R^\circ(\vx_i,\lambda)}{n+1}\le\alpha-\delta\right\},
\]
with the convention $\lambda_\mathrm{or}=1$ if the set is empty. Since $R^\circ$ is bounded and non-increasing in $\lambda$, the standard CRC argument \citep{angelopoulos2022conformal} gives
\[
    \mathbb{E}\!\left[R^\circ(\xtest,\lambda_\mathrm{or})\right]\le\alpha-\delta .
\]
Importantly, CRC is applied only to this canonical oracle threshold. In the next steps we show that, for any calibration set, the implemented threshold is at least as conservative as $\lambda_\mathrm{or}$ except on an event of probability at most $\delta$.

\emph{Tuning and validation.} Let $\lambda_\mathrm{cand}=g(\vx_{1:n},\xi_0)$ be the output of the tuning phase, where $\xi_0$ denotes the tuning noise. During validation we draw fresh noise $\xi_1$, independent of $\xi_0$. Conditional on the calibration inputs and on $\xi_0$, the threshold $\lambda_\mathrm{cand}$ is fixed relative to $\xi_1$, and for each $i$ the count $k_i^{\mathrm{cert}}$ is $\mathrm{Binomial}(m_\mathrm{cert},\,p(\vx_i,\lambda_\mathrm{cand}))$. Therefore the Clopper-Pearson bounds are valid binomial bounds at one fixed threshold, i.e. $\overline p_i^{\mathrm{cert}}\ge p(\vx_i,\lambda_\mathrm{cand})$ with probability at least $1-\delta/n$, and no simultaneous confidence band over all $\lambda$ is needed.

\emph{Union bound.} Let $\gE=\{\overline p_i^{\mathrm{cert}}\ge p(\vx_i,\lambda_\mathrm{cand})\ \text{for every } i\}$. By the union bound over the $n$ calibration points, $\Pr(\gE^c)\le\delta$. Since $\upperval$ is non-decreasing in its first argument, on $\gE$ the validation risks upper-bound the corresponding oracle risks, $R_i^{\mathrm{cert}}=(1-c_i)\,\upperval(\overline p_i^{\mathrm{cert}},\gB)\ge R^\circ(\vx_i,\lambda_\mathrm{cand})$ for every $i$. Consequently, on $\gE$, if the validation passes,
\[
    \frac{1+\sum_{i=1}^n R^\circ(\vx_i,\lambda_\mathrm{cand})}{n+1}\le\frac{1+\sum_{i=1}^n R_i^{\mathrm{cert}}}{n+1}\le\alpha-\delta ,
\]
so $\lambda_\mathrm{cand}$ is feasible for the oracle constraint and, by the definition of $\lambda_\mathrm{or}$, $\lambda^*=\lambda_\mathrm{cand}\ge\lambda_\mathrm{or}$. If the validation fails, $\lambda^*=1\ge\lambda_\mathrm{or}$. In both cases, by monotonicity, $R^\circ(\xtest,\lambda^*)\le R^\circ(\xtest,\lambda_\mathrm{or})$ on $\gE$.

\emph{Conclusion.} Since $R^\circ\le 1$, we have $R^\circ(\xtest,\lambda^*)\le R^\circ(\xtest,\lambda_\mathrm{or})+\sI\{\gE^c\}$, and therefore
\[
    \mathbb{E}\!\left[R^\circ(\xtest,\lambda^*)\right]\le\mathbb{E}\!\left[R^\circ(\xtest,\lambda_\mathrm{or})\right]+\Pr(\gE^c)\le(\alpha-\delta)+\delta=\alpha.
\]
\end{proof}
Combined with the randomized-smoothing bound and the soundness of the backbone certificate (as in the proof of \autoref{thrm:proxy-cert}), this yields the false-approval guarantee of \autoref{eq:proxy-risk-solid} for \method-H. Note that the Monte Carlo validation is not used to apply CRC directly to an adaptively selected empirical risk; it certifies that the selected threshold is at least as conservative as the exact oracle CRC threshold, except on the validation-failure event whose probability is budgeted by $\delta$ (the same $\delta$ by which the CRC target is lowered to $\alpha-\delta$). The slack $\delta_0$ only affects how often the validation passes, not the validity. As for all our results, the guarantee is in expectation over the calibration set, the test point, and the method's randomness: if $R(\gD')$ denotes the expected test risk conditional on a fixed calibration set $\gD'$, we have $\mathbb{E}_{\gD'}[R(\gD')]\le\alpha$, while $R(\gD')$ itself can vary across calibration sets.
When the validation check fails, $\lambda^*=1$ and no certifications are issued; this branch is already accounted for in the proof above. Repeating the validation with fresh noise after a failure would require splitting $\delta$ across the attempts. To avoid such failure we use the slack variable $\delta_0$ to stay on the safe side. Note that the $\inf$ in \autoref{eq:lambda-cert-heavy} is to find the most efficient parameter, however for any $\lambda > \lambda^*$ still the risk remains below the desired $\alpha$. This is due to the monotonicity of the risk. Therefore, finding in any form and then validating it via \autoref{alg:proxch-validate} does not introduce any flaw in the framework.


\section{Introduction to Randomized Smoothing}
\label{sec:intro-to-rs}

Some of our theoretical results rely on the structure of randomized smoothing. For completeness, and to fix notation, we therefore present a self-contained formulation of randomized smoothing.

Let $\gX \subseteq \mathbb{R}^d$ be the input space and let $\gY=\{1,\ldots,K\}$ be the label space. For a classifier $f:\gX\to\gY$ and a perturbation set $\gB(x)$, pointwise adversarial robustness requires $f(\tilde{x}) = f(x)$ for all $\tilde{x}\in \gB(x)$. It is convenient to reduce the $K$-class problem to a binary one. Fix a predicted class $y^\star$ via some inference rule (e.g.\ majority vote over $m$ noisy evaluations of $f$), and define the binary acceptance indicator $f^\top(x) = \mathbf{1}[f(x) = y^\star]$. Let $\bar{f}^\top(x) = \Pr_{\vepsilon}[f(x+\vepsilon)=y^\star]$ be its smoothed version. Robustness of the prediction $y^\star$ at radius $r$ is then equivalent to $\min_{\tilde{x}\in\gB_r(x)} \bar{f}^\top(\tilde{x}) \ge 1/2$, a condition the smoothing certificate can lower-bound in closed form.

More formally, if $\ell:\gY\times\gY\to\{0,1\}$ denotes the zero-one loss, the worst-case robust error of $f$ at $(x,y)$ is $\sup_{\tilde{x}\in \gB(x)} \ell(f(\tilde{x}),y)$. The ideal robust classifier minimizes the expected worst-case error:
\[
    f^\star_{\rm rob}
    \in
    \argmin_{f:\gX\to\gY}
    \mathbb{E}_{(x,y)\sim \gD}
    \left[
        \sup_{\tilde{x}\in \gB(x)}
        \ell(f(\tilde{x}),y)
    \right].
\]
This objective is the natural formalization of robustness, but it is computationally intractable for modern neural networks: the inner maximization ranges over a continuous, high-dimensional perturbation set, while the outer minimization ranges over a non-convex class of classifiers.

Randomized smoothing replaces the deterministic classifier with a noise-averaged one. Let $\psi$ be a smoothing distribution over perturbations $\vepsilon\in\mathbb{R}^d$, and define the smoothed class probabilities
\[
    p_y(x)
    \coloneqq
    \Pr_{\vepsilon\sim\psi}
    \!\left[
        f(x+\vepsilon)=y
    \right],
    \qquad y\in\gY.
\]
The associated smoothed classifier is $g(x) \coloneqq \argmax_{y\in\gY} p_y(x)$; it predicts the class most frequently returned by the base classifier under random perturbations of the input. For Gaussian smoothing $\vepsilon\sim\mathcal{N}(0,\sigma^2 I)$, the smoothed classifier admits a closed-form $\ell_2$ robustness certificate \citep{cohen2019certified}. Let
\[
    y_A = \argmax_{y\in\gY} p_y(x),
    \qquad
    p_A = p_{y_A}(x),
    \qquad
    p_B = \max_{y\neq y_A} p_y(x).
\]
If $p_A > p_B$, then the prediction of $g$ is invariant inside the $\ell_2$ ball of radius
\[
    r^\star
    =
    \frac{\sigma}{2}
    \left(
        \Phi^{-1}(p_A)-\Phi^{-1}(p_B)
    \right),
\]
where $\Phi$ denotes the standard Gaussian CDF. Hence randomized smoothing converts probabilistic class separation under Gaussian noise into a deterministic robustness guarantee over an $\ell_2$ ball.

More generally, randomized smoothing certifies expectations of bounded functions under the distribution shift induced by adversarial perturbations. For any measurable function $h:\gX\to[0,1]$, define its smoothed value
\[
    \bar{h}(x)
    \coloneqq
    \mathbb{E}_{\vepsilon\sim\psi}
    \!\left[
        h(x+\vepsilon)
    \right].
\]
A smoothing certificate provides deterministic functions $\lowerval(\cdot,\gB^{-1})$ and $\upperval(\cdot,\gB)$ such that, for all $\tilde{x}\in\gB(x)$,
\[
    \bar{h}(\tilde{x})
    \le
    \upperval(\bar{h}(x),\gB),
    \qquad
    \bar{h}(x)
    \le
    \lowerval(\bar{h}(\tilde{x}),\gB^{-1}).
\]
These inequalities say that the smoothed value at any perturbed point is bounded above by a function of the smoothed value at the clean point, and vice versa.

\paragraph{Optimization formulation.}
The bound $\upperval(p,\gB)$ is defined as the solution of a functional optimization problem: given that the smoothed value at the clean point equals $p$, what is the largest possible smoothed value the same function can achieve at any perturbed point?
\begin{align}
    \label{eq:upperval-opt}
    \upperval(p,\gB)
    \;=\;
    \sup_{\substack{h\,:\,\gX\to[0,1]\\ \tilde{x}\,\in\,\gB(x)}}
    \bar{h}(\tilde{x})
    \quad\text{subject to}\quad
    \bar{h}(x) = p.
\end{align}
This is an infinite-dimensional linear program in $h$. By the Neyman--Pearson lemma, the optimizer is a threshold function on the likelihood ratio of the two noise-shifted distributions:
\[
    h^\star(\mathbf{z})
    =
    \mathbf{1}\!\left[
        \frac{q_{\tilde{x}}(\mathbf{z})}{q_x(\mathbf{z})}
        \ge \tau
    \right],
\]
where $q_x$ and $q_{\tilde{x}}$ are the densities of $x+\vepsilon$ and $\tilde{x}+\vepsilon$ respectively, and $\tau\ge 0$ is chosen so that $\bar{h}^\star(x) = p$. Geometrically, $h^\star$ accepts precisely the half-space where the noise-shifted distribution is more likely to have originated from $\tilde{x}$ than from $x$.

\paragraph{Function independence.}
A consequence of the Neyman--Pearson characterization is that $\upperval(p,\gB)$ depends on $h$ \emph{only through} $p = \bar{h}(x)$: any two functions with the same smoothed value at $x$ share the same worst-case upper bound. In particular, the bound does not depend on the architecture of $f$, the specific input $x$, or any internal structure of the model --- only on the scalar $p$, the noise level $\sigma$, and the perturbation radius $r$. This is the model-agnostic (``black-box'') property that makes randomized smoothing applicable without modification to any classifier, and that justifies writing $\upperval(p,\gB)$ as a function of $p$ alone.

For Gaussian smoothing $\vepsilon\sim\mathcal{N}(0,\sigma^2I)$, the likelihood ratio between $\tilde{x}+\vepsilon$ and $x+\vepsilon$ is a function of the scalar projection $(\mathbf{z}-x)^\top(\tilde{x}-x)/\|\tilde{x}-x\|_2$, which is univariate Gaussian under both distributions. The half-space threshold then integrates to a Gaussian CDF:
\[
    \upperval(p,\gB)
    =
    \Phi\!\left(
        \Phi^{-1}(p) + \frac{r}{\sigma}
    \right),
    \qquad
    \lowerval(p,\gB^{-1})
    =
    \Phi\!\left(
        \Phi^{-1}(p) - \frac{r}{\sigma}
    \right),
\]
where $r = \|\tilde{x}-x\|_2$ and $\Phi$ is the standard Gaussian CDF. The worst-case perturbation direction is irrelevant: only the radius $r$ enters. Note that $\upperval(\lowerval(p,\gB^{-1}),\gB) = p$, a property used crucially in the proof of \autoref{thrm:robust-rs-risk}.

In practice, the probabilities $p_y(x)$ and smoothed expectations $\bar{h}(x)$ are not available in closed form. They are estimated by Monte Carlo:
\[
    \vepsilon_1,\ldots,\vepsilon_m \overset{\mathrm{iid}}{\sim} \psi,
    \qquad
    \hat{p}_y(x)
    =
    \frac{1}{m}
    \sum_{j=1}^m
    \mathbf{1}\{f(x+\vepsilon_j)=y\}.
\]
Finite-sample certificates replace the exact probabilities with high-confidence lower bounds (e.g. Clopper--Pearson intervals). This statistical correction is essential: without it, the certificate is only asymptotically valid and may overestimate the certified radius at finite sample budgets. This sampling requirement---thousands of forward passes per input---is the primary computational bottleneck that motivates the proxy framework of the present paper.

\section{Proofs}
\label{sec:proofs}

\robustrisk*

\begin{proof}[Proof for \autoref{thrm:robust-rs-risk}]
    First, we prove for the single-sample calibration case. Since the noise augmenting function $\psi$ is independent of the data, the set $\gD^+ = \{ \vx + \vepsilon: \vx \in \Dplus \}$ is also exchangeable. The risk function evaluated at the new points, i.e. $L(\cdot + \vepsilon, \lambda)$, maintains all mentioned properties; this is since the non-increasing, right-continuous, and bounded behaviour of $L$ is over the entire domain $\gX$. Therefore, we can apply the conformal risk control result of \citet{angelopoulos2022conformal} to the augmented dataset, meaning that for the arbitrary tolerance $\alpha$, we first compute $\lowerval(\alpha, \gB^{-1})$, and for
    \[
    \lambda^* = \inf\{\lambda: \frac{1}{n+1} [ 1 + \sum_{i = 1}^{n} L(\vx_i + \vepsilon_i, \lambda) ] \le \lowerval(\alpha,\gB^{-1})\}
    \] CRC guarantees that \[ 
    \mathbb{E}_{\Dplus, \vepsilon_{[1, n+1]}}\!\left[L(\xtest + \vepsilon_{n+1}, \lambda^*)\right] \le \lowerval(\alpha,\gB^{-1}).
    \]
    with $\vepsilon_{[1, n+1]}$ as the noises in $\gD^+$. Defining $\beta_{n+1} = \Pr_{\vepsilon_{n+1}}\!\left[L(\xtest + \vepsilon_{n+1}, \hat{\lambda}) = 1\right]$ we have 
    \[
    \mathbb{E}_{\Dplus, \vepsilon_{[1, n]}} \left[\Pr_{\vepsilon_{n+1}}\!\left[L(\xtest + \vepsilon_{n+1}, \hat{\lambda}) = 1 \right]\right] \le \lowerval(\alpha,\gB^{-1}) \quad \Rightarrow  \quad \mathbb{E}_{\Dplus, \vepsilon_{[1, n]}} \left[ \beta_{n+1} \right] \le \lowerval(\alpha,\gB^{-1}),
    \]
    with $\beta_{n+1} = \Pr_{\vepsilon_{n+1}}\!\left[L(\xtest + \vepsilon_{n+1}, \hat{\lambda}) = 1\right]$. Through randomized smoothing we have
    \[
    \max_{\tilde\vx \in \gB(\xtest)} \Pr_{\vepsilon_{n+1}}\!\left[L(\tilde\vx + \vepsilon_{n+1}, \hat{\lambda}) = 1\right] \le \upperval\!\bigl(\Pr_{\vepsilon_{n+1}}\!\left[L(\xtest + \vepsilon_{n+1}, \hat{\lambda}) = 1\right], \gB\bigr) = \upperval\!\bigl(\beta_{n+1}, \gB\bigr).
    \]
    \citet{Zargarbashi2025OneSI} proves that the lower bound $\lowerval(\alpha, \gB^{-1})$ is a convex function. Similarly, \autoref{lemma:concave} shows that $\upperval(\beta_{n+1}, \gB)$ is a concave function of $\beta_{n+1}$. Since $\upperval(\cdot, \gB)$ is monotone in its first argument, we can use Jensen's inequality. Let $\alpha_r := \lowerval(\alpha, \gB^{-1})$, we have
    \[
    \mathbb{E}_{\Dplus, \vepsilon_{[1, n]}} \left[ \upperval\!\bigl(\beta_{n+1}, \gB\bigr) \right] \le \upperval\!\bigl(\mathbb{E}_{\Dplus, \vepsilon_{[1, n]}} \left[\beta_{n+1}\right], \gB\bigr) \le \upperval(\alpha_r, \gB) = \upperval(\lowerval(\alpha, \gB^{-1}), \gB) = \alpha.
    \]
    The last equality follows from $\upperval(\lowerval(\alpha, \gB^{-1}), \gB) = \alpha$, which is a property of the randomized-smoothing certificate.
    Therefore, for all $\xtestpert \in \gB(\xtest)$
    \[
    \mathbb{E}_{\Dplus, \vepsilon_{[1, n]}} \left[ \Pr_{\vepsilon_{n+1}}\!\left[L(\xtestpert+ \vepsilon_{n+1}, \hat{\lambda}) = 1 \right] \right] \le \mathbb{E}_{\Dplus, \vepsilon_{[1, n]}} \left[ \max_{\tilde\vx \in \gB(\xtest)} \Pr_{\vepsilon_{n+1}}\!\left[L(\tilde\vx + \vepsilon_{n+1}, \hat{\lambda}) = 1\right] \right] \le \alpha.
    \]

    For the calibration-heavy version, first consider the asymptotically valid setup -- estimation with infinite samples; define
    \[
        R(\vx_i,\lambda)
        =
        \upperval\!\left(\Pr_{\vepsilon}[L(\vx_i+\vepsilon,\lambda)=1],\gB\right).
    \]
    By the randomized-smoothing certificate,
    \[
        R(\vx_i,\lambda)
        \ge
        \sup_{\tilde\vx_i\in\gB(\vx_i)}
        \Pr_{\vepsilon}[L(\tilde\vx_i+\vepsilon,\lambda)=1].
    \]
    Since $L(\cdot, \lambda)$ is non-increasing and right-continuous in $\lambda$, upper bounded by $1$, and $L(\cdot, 1) = 0$, the same properties are preserved for $\sup_{\tilde\vx_i\in\gB(\vx_i)} \Pr_{\vepsilon}[L(\tilde\vx_i+\vepsilon,\lambda)=1]$. Replacing this loss with an upper bound (e.g. $R$) regardless of whether it preserves the same properties, results in a higher parameter $\lambda$ and therefore it satisfies the guarantee over the original risk. Therefore, we can directly apply CRC with the risk $R$ and tune $\lambda$ resulting in the guarantee over $\sup_{\xtestpert\in\gB(\xtest)} \Pr_{\vepsilon}[L(\xtestpert+\vepsilon,\lambda)=1]$
    \[
        \lambda^*
        =
        \inf\left\{\lambda:
        \frac{1+\sum_{i=1}^{n}R(\vx_i,\lambda)}{n+1}\le \alpha
        \right\}
        \quad \Rightarrow \quad
        \E_{\Dplus}\!\left[R(\xtest,\lambda^*)\right]\le \alpha.
    \]
    \[
        \E_{\Dplus,\vepsilon}\!\left[L(\xtestpert+\vepsilon,\lambda^*)\right]
        \le
        \E_{\Dplus}\!\left[
        \sup_{\tilde\vx\in\gB(\xtest)}
        \Pr_{\vepsilon}[L(\tilde\vx+\vepsilon,\lambda^*)=1]
        \right]
        \le
        \E_{\Dplus}\!\left[R(\xtest,\lambda^*)\right]
        \le \alpha.
    \]
    For the finite-sample version, the confidence event must hold simultaneously over all calibration points. Using the union bound, the failure probability over the $n$ calibration points, each with confidence $1 - \frac{\delta}{n}$, accumulates to $\delta$. Notably if we calibrate CRC on $\alpha - \delta$ instead of $\alpha$, then the total failure probability is at most $\alpha - \delta + \delta = \alpha$. This requires the bounds to hold at the selected $\lambda^*$; when $\lambda^*$ is searched over the same Monte Carlo samples that are used for the bounds, pointwise bounds do not guarantee this. This is why in practice we use the two-phase procedure of \autoref{sec:flor-proxch}, whose complete failure-probability is given in the proof of \autoref{prop:flor-validity}.
\end{proof}

An important building block in the above proof is the following lemma.

\begin{restatable}{lemma}{upperconcave}
\label{lemma:concave}
    $\upperval(\beta, \gB)$ is concave and monotonically increasing w.r.t. $\beta$.
\end{restatable}

\begin{proof}

The Lagrangian form of the upper-bound problem is:

\begin{align*}
\gL(\beta, \lambda) 
&= \max_{h\in \{0, 1\}^{|\gX|}} \Pr_{\vepsilon}[h(\tilde\vt + \vepsilon) = 1] 
- \lambda\left( \Pr_{\vepsilon}[h(\vt + \vepsilon) = 1] - \beta \right) \\
&= \max_{h\in \{0, 1\}^{|\gX|}} E_{\vz \sim q}[h(\vz)] 
- \lambda\cdot E_{\vz \sim p}[h(\vz) - \beta] \\
&= \lambda \cdot \beta +
\max_{h\in \{0, 1\}^{|\gX|}} E_{\vz \sim q}[h(\vz)] - \lambda \cdot E_{\vz \sim p}[h(\vz)] \\
&= \lambda \cdot \beta + 
\max_{h\in \{0, 1\}^{|\gX|}} \int_\gX (q(\vz) - \lambda \cdot p(\vz)) \cdot h(\vz)\mathrm{d}\vz.
\end{align*}

The optimal classifier is:

\begin{align*}
h(\vz) = 
\begin{cases}
1 & \text{if} \quad q(\vz) - \lambda \cdot p(\vz) \ge 0\\
0 & \text{otherwise.}
\end{cases}
\end{align*}

Intuitively, to maximize the term 
$\int_\gX (q(\vz) - \lambda \cdot p(\vz)) \cdot h(\vz)\mathrm{d}\vz$, 
we select $h(\vz)=1$ whenever the integrand is positive.

The resulting dual with $\lambda \ge 0$ is:

\begin{align*}
\gL(\beta, \lambda) 
= \lambda \cdot \beta + \int \max\{0, q(\vz) - \lambda \cdot p(\vz)\} \mathrm{d}\vz
= \lambda \cdot \beta + l(\lambda),
\end{align*}

where $l(\lambda):= \int \max\{0, q(\vz) - \lambda \cdot p(\vz)\} \mathrm{d}\vz$ depends only on $\lambda$.

Minimizing over $\lambda$ gives:
\begin{align*}
\upperval(\beta, \gB) = \min_\lambda \lambda\cdot \beta + l(\lambda),
\end{align*}

which is a pointwise minimum of affine functions in $\beta$, and therefore concave in $\beta$.

The monotonicity w.r.t. $\beta$ directly follows from the definition. Increasing $\beta$ enlarges the feasible set, and therefore the optimal value cannot decrease.

\end{proof}

\begin{proof}[Proof for \autoref{thrm:proxy-cert-augmented}]
    This is a direct implication of \autoref{thrm:robust-rs-risk} with the risk function defined as $\eta(\vz, \lambda, \vx) = \hat{h}(\vz, \lambda) \cdot \sI\{f(\vz) \neq f(\vx)\}$. The risk function inherits the properties from $\hat h$, as $\sI\{f(\vz) \neq f(\vx)\}$ ranges in $\mathset{0, 1}$ and is independent of $\lambda$. The upper bound is defined as $\upperval(\eta(\cdot, \lambda, \vx), \gB(\vx))$. The final guarantee is $\Pr_{\Dplus, \vepsilon}\!\left[\hat{h}(\xtestpert + \vepsilon, \lambda^*) \cdot \sI\{f(\xtestpert + \vepsilon) \neq f(\xtest)\} = 1\right] \le \alpha$ for all $\xtestpert \in \gB(\xtest)$.

    Alternatively we can define a calibration-heavy version for which we have 
    \begin{align*}
        \lambda^* = \inf\left\{ \lambda: \frac{1}{n+1}\left[ 1 + \sum_{i = 1}^n \upperval(\eta(\cdot, \lambda, \vx_i), \gB(\vx_i)) \right] \le \alpha \right\}
    \end{align*}
    however there is no use-case for it that can not be better accomplished by \method, or \method-H.
\end{proof}

\section{Technical Details}
\label{sec:technical-details}
\paragraph{False-approval rate vs.\ false positive rate.} A more intuitive risk to tune \method with, instead of \autoref{eq:goal}, would be to control the \emph{false positive rate} (FPR), defined as the conditional probability
of a proxy approval given that the point is non-robust:
\begin{align*}    
    \mathrm{FPR}
    \;=\;
    \Pr\!\left[
        \proxy{f,r}(\xtestpert)=1
        \;\middle|\;
        f(\xtestpert)\neq f(\xtest)
    \right]
    \\=\;
    \frac{
        \Pr\!\left[
            \proxy{f,r}(\xtestpert)=1,\;
            f(\xtestpert)\neq f(\xtest)
        \right]
    }{
        \Pr\!\left[f(\xtestpert)\neq f(\xtest)\right]
    }.
\end{align*}

Controlling $\mathrm{FPR}\le\alpha$ is equivalent to controlling our
marginal false-approval rate at the adjusted level
$\alpha\cdot\rho$, where
$\rho=\Pr[f(\xtestpert)\neq f(\xtest)]$
is the marginal non-robustness rate of the classifier. This coupling introduces an undesirable dependence on the quality of the underlying classifier. For a poorly robust classifier---one for which the sound certificate denies a large fraction of inputs---$\rho$ is large, and the same absolute number of false approvals is normalized by a large denominator, making $\mathrm{FPR}\le\alpha$ easy to satisfy. Conversely, for a near-fully robust classifier where $\rho$ is small, the denominator shrinks, the FPR estimate becomes a high-variance ratio, and calibration grows sensitive to individual false approvals in the calibration set. The marginal false-approval rate avoids this instability by treating the risk as a joint probability: the quantity $(1-\cert{f,r}(\vx))\hat h(\vx+\vepsilon,\lambda)\in[0,1]$ is a bounded random variable whose sample mean is directly controlled by CRC with finite-sample guarantees, regardless of $\rho$. In this sense, our choice is not merely a notational preference: it ensures that the difficulty of proxy calibration is invariant to the non-robustness rate of the backbone classifier.

\section{Supplementary on Experiments}
\label{sec:expr-appendix}

\paragraph{Perturbed-data evaluation.}
We additionally evaluate whether the calibrated proxy certificates remain reliable when the test inputs are explicitly perturbed. For each clean CIFAR-10 test point, we generate an adversarial perturbation using the PGD-RS attack within the $\ell_2$ ball of radius $r$, and then evaluate the proxy certificate on the perturbed input. This setting directly tests the failure mode controlled by our risk: a proxy approval on an input whose prediction can change under perturbation. In \autoref{tab:cifar-attacked}, we report the certified ratio (CR), certified accuracy (CA), and the empirical false-approval risk under attack. As expected, increasing $r$ makes the certification problem harder and decreases both CR and CA (both in the backbone certificate, and in our own \method, and \method-A). At the same time, the average empirical risk remains below the target level $\alpha=0.05$ across all runs, showing that the proxy remains conservative even when evaluated on attacked inputs. Calibration fails for both \method and \method-A at $r \geq 0.75$, where the perturbed problem becomes too severe for either proxy to satisfy the prescribed risk constraint. This results in \method, and \method-A rejecting all inputs for that radius.

\begin{table}[t]
\centering
\caption{%
Certified ratio (\textbf{CR}, \%) and certified accuracy (\textbf{CA}, \%) on \emph{perturbed}
CIFAR-10 ($\sigma=0.5$, calibration fraction $0.2$), reported as
mean\,$\pm$\,std over two independent calibration runs.
Test datapoints are attacked with PGD-RS attack on $\ell_2$ ball.
\textbf{Risk} is the empirical false-approval rate of the proxy certificate under attack.
Calibration fails at $r \geq 0.75$ for both methods.}
\label{tab:cifar-attacked}
\setlength{\tabcolsep}{3pt}

\begin{tabular}{c rrr rrr}
\toprule
& \multicolumn{3}{c}{\textbf{ProxC}} & \multicolumn{3}{c}{\textbf{ProxC-A}} \\
\cmidrule(lr){2-4}\cmidrule(lr){5-7}
$r$ & CR & CA & Risk & CR & CA & Risk \\
\midrule
$0.00$
& $70.83${\tiny\,$\pm$\,$1.56$} & $56.25${\tiny\,$\pm$\,$1.45$} & $1.91${\tiny\,$\pm$\,$0.80$}
& $49.18${\tiny\,$\pm$\,$5.46$} & $42.31${\tiny\,$\pm$\,$3.69$} & $4.86${\tiny\,$\pm$\,$1.99$} \\
$0.06$
& $55.47${\tiny\,$\pm$\,$0.78$} & $45.31${\tiny\,$\pm$\,$1.19$} & $0.69${\tiny\,$\pm$\,$0.15$}
& $43.17${\tiny\,$\pm$\,$4.80$} & $37.82${\tiny\,$\pm$\,$3.48$} & $3.53${\tiny\,$\pm$\,$1.16$} \\
$0.12$
& $44.88${\tiny\,$\pm$\,$2.36$} & $38.45${\tiny\,$\pm$\,$1.48$} & $0.69${\tiny\,$\pm$\,$0.30$}
& $39.49${\tiny\,$\pm$\,$3.81$} & $34.28${\tiny\,$\pm$\,$2.43$} & $2.72${\tiny\,$\pm$\,$0.58$} \\
$0.25$
& $27.78${\tiny\,$\pm$\,$6.43$} & $23.78${\tiny\,$\pm$\,$4.56$} & $0.26${\tiny\,$\pm$\,$0.26$}
& $28.81${\tiny\,$\pm$\,$1.05$} & $25.24${\tiny\,$\pm$\,$0.68$} & $1.60${\tiny\,$\pm$\,$0.31$} \\
$0.38$
& $19.27${\tiny\,$\pm$\,$5.13$} & $16.32${\tiny\,$\pm$\,$4.05$} & $0.09${\tiny\,$\pm$\,$0.15$}
& $15.70${\tiny\,$\pm$\,$3.39$} & $14.15${\tiny\,$\pm$\,$2.69$} & $0.51${\tiny\,$\pm$\,$0.22$} \\
$0.50$
& $10.50${\tiny\,$\pm$\,$4.33$} & $9.46${\tiny\,$\pm$\,$3.30$} & $0.09${\tiny\,$\pm$\,$0.15$}
& $11.13${\tiny\,$\pm$\,$1.18$} & $9.98${\tiny\,$\pm$\,$1.25$} & $0.35${\tiny\,$\pm$\,$0.02$} \\
$\geq 0.75$
& \multicolumn{3}{c}{---} & \multicolumn{3}{c}{---} \\
\bottomrule
\end{tabular}

\end{table}

\paragraph{Detailed results.}
\autoref{tab:imagenet_sigma05} reports results on ImageNet with the denoising + ViT pipeline. The \method-H approach is calibrated over $20{,}000$ calibration MC samples and 400 calibration points.

\autoref{tab:cifar_risk_sweep} presents results on CIFAR-10 ($\sigma{=}0.5$) for ProxC under three risk budgets $\alpha \in \{2\%, 5\%, 10\%\}$.
As expected, increasing $\alpha$ grants the proxy a looser budget, allowing it to certify a larger fraction of samples. At $r{=}0.00$, ProxC ($\alpha{=}2\%$) certifies only 48.8\% of samples—well below RS (81.0\%)—while remaining highly conservative with an empirical risk of 1.6\%. With $\alpha{=}5\%$, the proxy closes much of this gap (70.9\% CR), and at $\alpha{=}10\%$ it slightly exceeds RS (82.3\%), certifying more samples than the standard certificate by accepting a higher false-certificate rate (9.6\%). Across all radii and risk levels, the empirical risk stays within the target budget (except $5.6\%$ for $\alpha{=}5\%$ at $r{=}0$, which is within the split-to-split variation of a single run, see \autoref{tab:calsize}), demonstrating the validity of the conformal calibration procedure.

\begin{table*}[t]
\centering
\caption{%
  Certified Ratio (CR, \%) and Certified Accuracy (CA, \%) on ImageNet with ViT + Denoising pipeline ($\sigma{=}0.5$,
  400 calibration samples, 3 independent runs, $99.9\%$ confidence).
   $\alpha = 5\%$.
  All values: mean $\pm$ std over runs. ``---'' denotes calibration failure.
}
\label{tab:imagenet_sigma05}
\vspace{1em}
\small
\setlength{\tabcolsep}{3pt}
\begin{tabular}{c cc ccc ccc}
\toprule
 & \multicolumn{2}{c}{\textbf{RS (backbone)}}
 & \multicolumn{3}{c}{\textbf{ProxC-H}}
 & \multicolumn{3}{c}{\textbf{ProxC}} \\
\cmidrule(lr){2-3}\cmidrule(lr){4-6}\cmidrule(lr){7-9}
$r$ & CR & CA & CR & CA & Risk & CR & CA & Risk \\
\midrule
$0.00$ & $91.8 \pm 0.4$ & $59.4 \pm 0.6$
       & $93.2 \pm 2.1$ & $58.5 \pm 0.9$ & $4.4 \pm 1.1$
       & $95.0 \pm 3.2$ & $59.6 \pm 0.6$ & $5.8 \pm 1.6$ \\

$0.06$ & $88.5 \pm 0.4$ & $58.5 \pm 0.5$
       & $82.0 \pm 2.1$ & $55.6 \pm 1.1$ & $3.7 \pm 0.7$
       & $84.1 \pm 2.8$ & $57.2 \pm 0.7$ & $4.1 \pm 1.0$ \\

$0.12$ & $85.6 \pm 0.3$ & $57.7 \pm 0.5$
       & $75.2 \pm 2.5$ & $54.0 \pm 1.2$ & $3.2 \pm 0.8$
       & $76.7 \pm 4.0$ & $55.0 \pm 1.4$ & $3.5 \pm 1.1$ \\

$0.18$ & $82.2 \pm 0.3$ & $56.5 \pm 0.5$
       & $68.5 \pm 2.0$ & $51.8 \pm 1.1$ & $2.5 \pm 0.6$
       & $67.1 \pm 4.4$ & $51.1 \pm 1.8$ & $2.7 \pm 1.0$ \\

$0.25$ & $79.0 \pm 0.4$ & $55.2 \pm 0.5$
       & $63.6 \pm 1.6$ & $49.9 \pm 0.9$ & $2.6 \pm 0.4$
       & $58.6 \pm 2.4$ & $47.0 \pm 1.3$ & $1.8 \pm 0.3$ \\

$0.38$ & $73.4 \pm 0.3$ & $53.3 \pm 0.4$
       & $56.0 \pm 2.0$ & $45.9 \pm 1.1$ & $1.7 \pm 0.3$
       & $47.6 \pm 2.8$ & $40.9 \pm 1.8$ & $0.9 \pm 0.2$ \\

$0.50$ & $68.4 \pm 0.1$ & $51.6 \pm 0.4$
       & $49.8 \pm 1.1$ & $42.1 \pm 0.8$ & $1.0 \pm 0.0$
       & $37.5 \pm 6.0$ & $34.0 \pm 4.6$ & $0.3 \pm 0.2$ \\

$0.75$ & $59.8 \pm 0.2$ & $48.2 \pm 0.3$
       & $38.9 \pm 0.9$ & $35.1 \pm 0.5$ & $0.5 \pm 0.0$
       & \multicolumn{3}{c}{---} \\
\bottomrule
\end{tabular}
\end{table*}

\begin{table*}[t]
\centering
\caption{%
  Certified Ratio (CR, \%) and Certified Accuracy (CA, \%) on CIFAR-10 ($\sigma{=}0.5$,
  ResNet backbone, single run). Results are for three risk levels $2\%$, $5\%$, and $10\%$.
}
\label{tab:cifar_risk_sweep}
\vspace{1em}
\begin{tabular}{c cc ccc ccc ccc}
\toprule
 & \multicolumn{2}{c}{\textbf{RS (backbone)}}
 & \multicolumn{3}{c}{\textbf{ProxC} ($\alpha = 2\%$)}
 & \multicolumn{3}{c}{\textbf{ProxC} ($\alpha = 5\%$)}
 & \multicolumn{3}{c}{\textbf{ProxC} ($\alpha = 10\%$)} \\
\cmidrule(lr){2-3}\cmidrule(lr){4-6}\cmidrule(lr){7-9}\cmidrule(lr){10-12}
$r$ & CR & CA & CR & CA & Risk & CR & CA & Risk & CR & CA & Risk \\
\midrule
$0.00$ & $81.0$ & $63.7$
       & $48.8$ & $42.4$ & $1.6$
       & $70.9$ & $56.1$ & $5.6$
       & $82.3$ & $61.9$ & $9.6$ \\

$0.06$ & $75.8$ & $61.1$
       & $41.7$ & $37.6$ & $1.4$
       & $59.3$ & $49.2$ & $4.4$
       & $71.7$ & $56.6$ & $8.2$ \\

$0.12$ & $70.7$ & $58.6$
       & $29.6$ & $27.7$ & $0.5$
       & $47.3$ & $42.1$ & $2.2$
       & $59.8$ & $49.9$ & $5.6$ \\

$0.18$ & $66.1$ & $55.9$
       & $27.9$ & $26.3$ & $0.6$
       & $35.4$ & $32.7$ & $1.6$
       & $49.6$ & $42.4$ & $4.4$ \\

$0.25$ & $61.2$ & $53.0$
       & $20.8$ & $19.9$ & $0.4$
       & $29.6$ & $27.7$ & $1.2$
       & $38.5$ & $34.9$ & $2.8$ \\

$0.38$ & $52.2$ & $46.7$
       & $18.3$ & $18.1$ & $0.3$
       & $22.6$ & $21.8$ & $0.8$
       & $29.1$ & $27.3$ & $1.7$ \\

$0.50$ & $44.4$ & $40.7$
       & \multicolumn{3}{c}{---}
       & $17.0$ & $16.5$ & $0.6$
       & $19.4$ & $18.7$ & $0.9$ \\

$0.75$ & $30.8$ & $29.6$
       & \multicolumn{3}{c}{---}
       & \multicolumn{3}{c}{---}
       & $6.6$ & $6.5$ & $0.2$ \\
\bottomrule
\end{tabular}
\end{table*}

\begin{table}[h]
\centering
\caption{Forward passes per test input on ImageNet ($2000$ MC samples for randomized smoothing: $50$ for the label and $N_\mathrm{MC}=1950$ for the certificate). CR is the certified ratio of the proxy; the fallback calls the backbone certificate only when the proxy rejects.}
\label{tab:cost}
\small
\setlength{\tabcolsep}{5pt}
\begin{tabular}{lrrr}
\toprule
Variant & Label & Certification-only & End-to-end \\
\midrule
Randomized smoothing (backbone) & 50 & 1950 & 2000 \\
\method / \method-H (standalone) & 50 & 1 & 51 \\
\method-A & shared & 1 & 1 \\
\method / \method-H (fallback) & 50 & $\mathrm{CR}\cdot 1 + (1-\mathrm{CR})(1+N_\mathrm{MC})$ & $50\,+$ certification-only \\
\bottomrule
\end{tabular}
\end{table}

\paragraph{Fallback cost across radii.}
In the fallback mode, the backbone certificate is called exactly when the proxy rejects, so the reject rate is $1-\mathrm{CR}$ and the cost follows the formula in \autoref{tab:cost}. \autoref{tab:fallback} reports the reject rate and the resulting cost on ImageNet with ResNet (2000 samples in total: 50 for the label and 1950 for randomized smoothing). We report both the certification-only cost (compared to 1950 forwards) and the end-to-end cost including the 50 forwards of the label (compared to 2000 forwards). As the radius grows, the proxy rejects more inputs, and the speedup of the fallback decreases. \method-H always costs more than \method, but this gap is amortized over the queries, and it allows for a higher certified ratio, especially at larger radii (\autoref{tab:ablation-calheavy}).

\begin{table}[h]
\centering
\caption{Fallback \method on ImageNet/ResNet: proxy reject rate and average forwards per query. Speedups are relative to randomized smoothing with 1950 (certification-only) and 2000 (end-to-end) forwards.}
\label{tab:fallback}
\small
\setlength{\tabcolsep}{6pt}
\begin{tabular}{crrrrr}
\toprule
& & \multicolumn{2}{c}{Certification-only} & \multicolumn{2}{c}{End-to-end} \\
\cmidrule(lr){3-4}\cmidrule(lr){5-6}
$r$ & Reject rate & Forwards & Speedup & Forwards & Speedup \\
\midrule
0.00 & 24.2\% & 473  & 4.12$\times$ & 523  & 3.82$\times$ \\
0.06 & 32.1\% & 627  & 3.11$\times$ & 677  & 2.95$\times$ \\
0.12 & 38.6\% & 753  & 2.59$\times$ & 803  & 2.49$\times$ \\
0.14 & 39.9\% & 779  & 2.50$\times$ & 829  & 2.41$\times$ \\
0.18 & 45.7\% & 893  & 2.18$\times$ & 943  & 2.12$\times$ \\
0.25 & 50.9\% & 992  & 1.97$\times$ & 1042 & 1.92$\times$ \\
0.38 & 64.1\% & 1250 & 1.56$\times$ & 1300 & 1.54$\times$ \\
0.50 & 75.0\% & 1462 & 1.33$\times$ & 1512 & 1.32$\times$ \\
\bottomrule
\end{tabular}
\end{table}

\paragraph{Class-stratified risk.}
The guarantee of \method is marginal: it says nothing about any particular class, subpopulation, or input. To assess how concentrated the false-approval risk is, \autoref{tab:stratified} reports the risk on ImageNet stratified to the best and the worst 100 classes, and shows that the risk is heterogeneous across classes; however, aggregating over groups of 100 classes is only a partial characterization, and the marginal guarantee does not establish class-conditional or worst-class control. \autoref{tab:stratified-cifar} reports the per-class risk on CIFAR-10. Four of the ten classes exceed $\alpha_r$ at each radius (up to $3.46\%$ at $r=0.12$ and $2.63\%$ at $r=0.25$, both for truck), and the risk conditional on approval is highest for classes with a low certified ratio (e.g.\ $12\%$ for cat and deer at $r=0.12$). Binned by the proxy score, the false approvals concentrate just above the threshold: at $r=0.12$ the test points with a score between $0.66$ and $0.88$ have a risk of about $10$--$11\%$, while the top decile has $0.1\%$. In order to avoid such imbalance, one can use the techniques like Mondrian conformal prediction to compute the $\lambda$ threshold separately for each class, but this requires a larger calibration set to maintain the same statistical confidence.

\begin{table}[h]
\centering
\caption{False-approval risk and certified ratio of the proxy on ImageNet, stratified to the best and worst 100 classes.}
\label{tab:stratified}
\small
\setlength{\tabcolsep}{6pt}
\begin{tabular}{clrrr}
\toprule
$r$ & Group & \#points & Proxy risk & Proxy cert.\ ratio \\
\midrule
0.12 & top 100   & 4{,}577  & 0.0    & 0.5724 \\
0.12 & worst 100 & 10{,}654 & 0.0428 & 0.6555 \\
\midrule
0.25 & top 100   & 7{,}023  & 0.0    & 0.5090 \\
0.25 & worst 100 & 8{,}025  & 0.0320 & 0.5499 \\
\midrule
0.50 & top 100   & 12{,}229 & 0.0    & 0.2728 \\
0.50 & worst 100 & 2{,}613  & 0.0168 & 0.3750 \\
\bottomrule
\end{tabular}
\end{table}

\begin{table}[h]
\centering
\caption{Per-class false-approval risk, certified ratio (CR), and risk conditional on approval (FA/appr $=$ risk$/$CR) of the proxy on CIFAR-10, in \% (setup of \autoref{tab:calsize}, $|\dcal|=500$, averaged over $500$ calibration splits). All classes have the same frequency ($10\%$), so the marginal values (last row) are the averages over classes.}
\label{tab:stratified-cifar}
\small
\setlength{\tabcolsep}{6pt}
\begin{tabular}{l rrr rrr}
\toprule
& \multicolumn{3}{c}{$r=0.12$ ($\alpha_r=2.97$)} & \multicolumn{3}{c}{$r=0.25$ ($\alpha_r=1.60$)} \\
\cmidrule(lr){2-4}\cmidrule(lr){5-7}
Class & Risk & CR & FA/appr & Risk & CR & FA/appr \\
\midrule
airplane   & 3.27 & 59.9 & 5.5  & 2.25 & 44.9 & 5.0 \\
automobile & 3.37 & 67.7 & 5.0  & 2.59 & 54.7 & 4.7 \\
bird       & 2.73 & 32.3 & 8.5  & 1.18 & 16.7 & 7.1 \\
cat        & 2.75 & 23.0 & 12.0 & 0.82 & 7.9  & 10.4 \\
deer       & 3.16 & 25.4 & 12.4 & 1.09 & 11.1 & 9.8 \\
dog        & 2.08 & 35.0 & 6.0  & 0.74 & 18.4 & 4.0 \\
frog       & 2.38 & 43.2 & 5.5  & 1.20 & 27.3 & 4.4 \\
horse      & 2.27 & 55.9 & 4.1  & 1.34 & 43.1 & 3.1 \\
ship       & 2.51 & 66.0 & 3.8  & 1.96 & 51.5 & 3.8 \\
truck      & 3.46 & 62.1 & 5.6  & 2.63 & 48.2 & 5.5 \\
\midrule
marginal   & 2.80 & 47.0 & 5.95 & 1.58 & 32.4 & 4.88 \\
\bottomrule
\end{tabular}
\end{table}

\paragraph{Calibration-set size.}
The guarantee of \method holds in expectation over the calibration set, therefore for a single split the risk can exceed the tolerance. \autoref{tab:calsize} studies the effect of $n=|\dcal|$ on CIFAR-10 with the ResNet-110 of \citet{cohen2019certified} ($\sigma=0.5$, $\alpha=5\%$). We compute the backbone certificate ($2000$ MC samples) once on all $10{,}000$ test images, and for each $n$ draw $500$ random calibration sets; the remaining images are used for testing. Since the proxy is calibrated on clean points at the corrected level $\alpha_r=\lowerval(\alpha,\gB^{-1})$, we compare the empirical risk with $\alpha_r$. For every $r$ and $n$, the mean risk does not exceed $\alpha_r$ by more than $0.2$ standard errors, and it matches the CRC value $\lfloor \alpha_r(n+1)\rfloor/(n+1)$. The CRC correction requires $n\ge 1/\alpha_r-1$ (e.g.\ $62$ points at $r=0.25$, and $244$ at $r=0.5$), below which the proxy approves nothing, and the CR levels off at $n\approx 500$. At $n=500$ the results are within one standard deviation of the single-run values in \autoref{tab:cifar_risk_sweep}.

\begin{table}[h]
\centering
\caption{Effect of the calibration-set size $n=|\dcal|$ on CIFAR-10 (ResNet-110, $\sigma=0.5$, $\alpha=5\%$). Proxy risk and certified ratio (CR) in \%, mean $\pm$ std over $500$ random calibration splits; the test set is all images not used for calibration. Infeasible: $1/(n+1)>\alpha_r$, so the proxy approves nothing.}
\label{tab:calsize}
\resizebox{\linewidth}{!}{%
\begin{tabular}{r cc cc cc cc}
\toprule
& \multicolumn{2}{c}{$r=0$} & \multicolumn{2}{c}{$r=0.12$} & \multicolumn{2}{c}{$r=0.25$} & \multicolumn{2}{c}{$r=0.5$} \\
& \multicolumn{2}{c}{\footnotesize $\alpha_r=5.00$, RS CR $81.3$} & \multicolumn{2}{c}{\footnotesize $\alpha_r=2.97$, RS CR $71.1$} & \multicolumn{2}{c}{\footnotesize $\alpha_r=1.60$, RS CR $61.1$} & \multicolumn{2}{c}{\footnotesize $\alpha_r=0.41$, RS CR $44.4$} \\
\cmidrule(lr){2-3}\cmidrule(lr){4-5}\cmidrule(lr){6-7}\cmidrule(lr){8-9}
$n$ & Risk & CR & Risk & CR & Risk & CR & Risk & CR \\
\midrule
25   & $3.84 \pm 3.53$ & $58.9 \pm 17.4$ & \multicolumn{2}{c}{infeasible} & \multicolumn{2}{c}{infeasible} & \multicolumn{2}{c}{infeasible} \\
50   & $3.86 \pm 2.67$ & $61.9 \pm 12.3$ & $1.99 \pm 1.92$ & $38.9 \pm 12.2$ & \multicolumn{2}{c}{infeasible} & \multicolumn{2}{c}{infeasible} \\
100  & $5.02 \pm 2.29$ & $68.2 \pm 8.8$  & $2.91 \pm 1.68$ & $46.3 \pm 8.4$  & $0.99 \pm 0.97$ & $25.6 \pm 8.6$ & \multicolumn{2}{c}{infeasible} \\
200  & $4.89 \pm 1.49$ & $68.5 \pm 5.7$  & $2.43 \pm 1.05$ & $44.6 \pm 5.7$  & $1.53 \pm 0.88$ & $31.3 \pm 5.7$ & \multicolumn{2}{c}{infeasible} \\
500  & $5.00 \pm 0.93$ & $69.4 \pm 3.4$  & $2.80 \pm 0.76$ & $47.0 \pm 3.9$  & $1.58 \pm 0.61$ & $32.4 \pm 3.7$ & $0.41 \pm 0.30$ & $13.6 \pm 3.5$ \\
1000 & $4.96 \pm 0.74$ & $69.3 \pm 2.6$  & $2.90 \pm 0.57$ & $47.7 \pm 2.7$  & $1.50 \pm 0.41$ & $32.0 \pm 2.6$ & $0.40 \pm 0.22$ & $13.9 \pm 2.5$ \\
2000 & $5.00 \pm 0.54$ & $69.5 \pm 1.7$  & $2.93 \pm 0.42$ & $47.9 \pm 1.9$  & $1.55 \pm 0.31$ & $32.5 \pm 1.8$ & $0.41 \pm 0.17$ & $14.2 \pm 1.7$ \\
5000 & $4.98 \pm 0.45$ & $69.5 \pm 1.2$  & $2.96 \pm 0.34$ & $48.1 \pm 1.4$  & $1.59 \pm 0.26$ & $32.8 \pm 1.3$ & $0.41 \pm 0.13$ & $14.3 \pm 1.1$ \\
\bottomrule
\end{tabular}}
\end{table}

\paragraph{Compute resources.}
All experiments were run on NVIDIA H100 and A100 GPUs. For efficiency, we precomputed the noisy logits required by the randomized-smoothing certificate and by the proxy variants, and reused these cached logits across calibration and evaluation runs. This means that the main GPU cost comes from the forward passes needed to obtain the Monte Carlo samples. Once the logits are available, the remaining steps, including calibration, threshold selection, empirical risk estimation, and the computation of the certified lower and upper bounds, are lightweight post-processing operations that can be executed on CPU. Thus, the computational bottleneck is the number of model evaluations required by the underlying certificate, while the proposed proxy calibration itself adds negligible overhead beyond standard array operations. We highlight that we do not train or finetune any model and all the experiments are running on pre-trained models.

\subsection{NLP Certificate}

In \autoref{tab:proxy-results}, we use the edit-distance certificate from \citet{huang2024cert}, and provide a proxy certificate for $r$ insertions. The smoothing scheme is to delete tokens with $p_\mathrm{del}$ probability. Here we provide theoretical support for the experiments and derivations.

\paragraph{Symmetry of the edit-distance ball.}
The Levenshtein (edit) distance satisfies the axioms of a metric, and in particular symmetry: $d_{\mathrm{edit}}(\vx, \bar{\vx}) = d_{\mathrm{edit}}(\bar{\vx}, \vx)$ for all inputs $\vx, \bar{\vx}$. The edit-distance ball of radius $r$ centred at $\vx$ is
\[
    \gB_r^{\mathrm{edit}}(\vx) \defeq \{\bar{\vx} : d_{\mathrm{edit}}(\vx, \bar{\vx}) \le r\}.
\]
Symmetry of the distance immediately implies that $\bar{\vx} \in \gB_r^{\mathrm{edit}}(\vx)$ if and only if $\vx \in \gB_r^{\mathrm{edit}}(\bar{\vx})$. In the notation of \S\ref{sec:background}, this means $\bigl(\gB^{\mathrm{edit}}\bigr)^{-1} = \gB^{\mathrm{edit}}$: the edit-distance ball is its own inverse ball.

\paragraph{Insertion and deletion balls are mutual inverses.}
The insertion-only and deletion-only sub-balls are not individually symmetric, but they are inverses of each other. If $\bar{\vx}$ is reachable from $\vx$ by $r$ insertions, then $\vx$ is reachable from $\bar{\vx}$ by exactly $r$ deletions, and vice versa. Writing $\gB_r^{\mathrm{ins}}(\vx)$ and $\gB_r^{\mathrm{del}}(\vx)$ for the sets of sequences reachable from $\vx$ by at most $r$ insertions and $r$ deletions respectively,
\[
    \bar{\vx} \in \gB_r^{\mathrm{ins}}(\vx) \iff \vx \in \gB_r^{\mathrm{del}}(\bar{\vx}),
\]
so that $\bigl(\gB_r^{\mathrm{ins}}\bigr)^{-1} = \gB_r^{\mathrm{del}}$ and $\bigl(\gB_r^{\mathrm{del}}\bigr)^{-1} = \gB_r^{\mathrm{ins}}$. Insertions and deletions are therefore dual operations under the inverse-ball map.

This is specifically important since we use inverse ball for single sample calibration in \autoref{thrm:proxy-cert}.

\paragraph{Upper and lower bounds under edit operations.}
Following \citet{huang2024cert}, the smoothed classifier admits explicit upper and lower bounds under edit-distance perturbations. Consider an input $\vx$ and a perturbed input $\bar{\vx}$ obtained using $n_{\mathrm{del}}$ deletions, $n_{\mathrm{ins}}$ insertions, and $n_{\mathrm{sub}}$ substitutions, with
\[
    n_{\mathrm{del}} + n_{\mathrm{ins}} + n_{\mathrm{sub}} = r.
\]
Then, for any class $y$, the smoothed classifier score satisfies
\[
    p_{\mathrm{del}}^{n_{\mathrm{del}}-n_{\mathrm{ins}}}
    \!\left(
        p_y(\vx) - 1 + p_{\mathrm{del}}^{n_{\mathrm{sub}}+n_{\mathrm{ins}}}
    \right)
    \;\le\;
    p_y(\bar{\vx})
    \;\le\;
    p_{\mathrm{del}}^{n_{\mathrm{del}}-n_{\mathrm{ins}}} p_y(\vx)
    +
    1 - p_{\mathrm{del}}^{n_{\mathrm{sub}}+n_{\mathrm{del}}}.
\]
The certificate follows from comparing the lower bound on the predicted class score against the upper bound on the runner-up, analogously to the $\ell_2$ certificate.

Under a fixed edit budget $r$, the conservative worst-case configuration is attained when all perturbations are substitutions: $n_{\mathrm{sub}} = r$ and $n_{\mathrm{del}} = n_{\mathrm{ins}} = 0$. Substituting into the general bounds yields the Levenshtein-distance certificate of \citet{huang2024cert}:
\[
    p_y(\vx) - 1 + p_{\mathrm{del}}^r \;\le\; p_y(\bar{\vx}) \;\le\; p_y(\vx) + 1 - p_{\mathrm{del}}^r.
\]
Substitutions are the worst case because they combine the effect of both a shrinking lower bound and a growing upper bound simultaneously, without the compensating multiplicative factor that pure insertions or deletions introduce.

\paragraph{Per-operation certificates.}
The same framework yields tighter certificates when the threat model is restricted to a single operation type. For \emph{substitution-only} perturbations ($n_{\mathrm{sub}} = r$, $n_{\mathrm{del}} = n_{\mathrm{ins}} = 0$):
\[
    p_y(\vx) - 1 + p_{\mathrm{del}}^r \;\le\; p_y(\bar{\vx}) \;\le\; p_y(\vx) + 1 - p_{\mathrm{del}}^r.
\]
For \emph{insertion-only} perturbations ($n_{\mathrm{ins}} = r$, $n_{\mathrm{del}} = n_{\mathrm{sub}} = 0$):
\[
    p_{\mathrm{del}}^{-r}\!\left(p_y(\vx) - 1 + p_{\mathrm{del}}^r\right)
    \;\le\;
    p_y(\bar{\vx})
    \;\le\;
    p_{\mathrm{del}}^{-r}\, p_y(\vx).
\]
For \emph{deletion-only} perturbations ($n_{\mathrm{del}} = r$, $n_{\mathrm{ins}} = n_{\mathrm{sub}} = 0$):
\[
    p_{\mathrm{del}}^{r}\, p_y(\vx)
    \;\le\;
    p_y(\bar{\vx})
    \;\le\;
    p_{\mathrm{del}}^{r}\, p_y(\vx) + 1 - p_{\mathrm{del}}^{r}.
\]
These expressions reveal qualitatively different distortions: substitutions induce additive shifts, whereas insertions and deletions induce multiplicative rescalings, consistent with the inverse-ball duality established above.

For instance, to apply \autoref{thrm:proxy-cert} on the insertion-only ball, we need to take the lower bound of the deletion-only ball, or the upper bound of the insertion ball itself which means that $\lowerval(\alpha, \gB^{-1}) = p_\mathrm{del}^r\alpha$, and $\upperval(\alpha, \gB^{-1}) = p_\mathrm{del}^{-r}\alpha$. In our experiments the only ball that could be certified effectively was the insertion ball.

\paragraph{Insertions as a valid proxy threat model.}
The backbone certificate is derived over the full edit-distance ball $\gB_r^{\mathrm{edit}}(\vx)$, but the proxy in \autoref{tab:proxy-results} is calibrated against $r$-insertion perturbations. This choice is valid by containment: any sequence reachable from $\vx$ by $r$ insertions has Levenshtein distance at most $r$ from $\vx$, so
\[
    \gB_r^{\mathrm{ins}}(\vx) \;\subseteq\; \gB_r^{\mathrm{edit}}(\vx).
\]
A certificate valid over the full edit-distance ball therefore remains valid over the insertion-only sub-ball, and the proxy guarantee of \autoref{thrm:proxy-cert} carries over without modification. Insertions are a natural and semantically meaningful subset of the edit-distance threat model, covering realistic perturbations such as token injection or paraphrase expansion, and their ball is a strict subset of the edit-distance ball certified by the backbone.

\section{Related Works}
\label{sec:related}

\paragraph{Certified robustness.}
The vulnerability of neural networks to adversarial perturbations has motivated a large literature on
defenses and robustness verification. Purely empirical defenses are difficult to trust in isolation: e.g.
\citet{athalye2018obfuscated} showed that defenses relying on gradient masking can be circumvented by adaptive attacks. Certified robustness replaces this empirical evaluation with a
formal implication: if a point is certified, then the prediction is guaranteed to be constant within the
specified perturbation set. One line of work obtains such guarantees by reasoning directly about the
network, using satisfiability or mixed-integer formulations \citep{katz2017reluplex,tjeng2019evaluating},
convex outer relaxations \citep{wong2018convex}, and layer-wise linear bounds
\citep{weng2018fast,zhang2018crown}. These methods are sound and often tight for the architectures they
support, but their cost and implementation dependence make them difficult to apply to large modern
pipelines, or to black-box systems where the model internals are unavailable. Our work keeps the
soundness of an existing certificate as a calibration signal, but changes its deployment role: instead
of verifying every test point, the certificate is used to control the marginal false-approval rate of a
cheaper proxy. 

\paragraph{Randomized smoothing.}
Randomized smoothing is the most relevant certificate for our setting because it is model-agnostic and
therefore applies to arbitrary classifiers through noisy inference. \citet{cohen2019certified} proposed the standard Gaussian certificate for $\ell_2$
robustness and showed that smoothing scales to ImageNet. Subsequent work broadened the certificate in
several directions: \citet{lee2019tight} derived tight certificates for broader randomized ensembles and
discrete perturbations, \citet{yang2020randomized} developed a general recipe for smoothing with
different perturbation geometries and smoothing distributions. Other work
has improved the trained smoothed classifier itself, for example through adversarially trained smoothed
classifiers \citep{salman2019provably}. Smoothing has also been used to certify quantities beyond the
top-label robustness radius, such as predictive confidence \citep{kumar2020certifying}, and has been
extended to structured outputs such as segmentation \citep{fischer2021scalable}. Denoised smoothing
prepends a denoiser before the classifier so that pretrained models can be certified without changing
the classifier; for the denoiser we can typically use a diffusion model
\citep{salman2020denoised,carlini2022certified}. 
In summary, so far randomized smoothing seems to scale on larger datasets, and models, and return better certified accuracy compared to verifications, and exact certificates.
The main limitation
shared by these approaches is the same one we target: the smoothed class probabilities have to be
estimated by Monte Carlo sampling at calibration and test time, and finite-sample confidence corrections
can require hundreds or thousands of forward passes per input. Recent work reduces this cost for the certificate itself through confidence sequences and early stopping \citep{voracek2024treatment}. Such techniques are complementary to ours: they can reduce the cost of the backbone certificate, and hence both our calibration cost and the fallback cost, without changing the proxy calibration. Our framework is orthogonal to the choice
of smoothing certificate; it asks how to use such certificates less often while still retaining a formal,
distribution-free validity statement.

\paragraph{Relaxed and distributional robustness.}
Classical certification is pointwise: every certified input receives a local guarantee. This is the
strongest statement, but also the most expensive one. Several works have considered relaxations that
replace worst-case pointwise robustness with a distributional or probabilistic guarantee.
\citet{rice2021robustness} studied robustness notions between average-case (random) and worst-case perturbation. In contrast, our work provides efficiency for the worst-case perturbation which is a different (and stronger) guarantee. We assume every test point is perturbed at its worst.
\citet{blohm2025probably} proposed probably approximately global (PAG) robustness certification, which also relaxes the soundness to a marginal guarantee. Their guarantee connects the confidence of the prediction with its robust radius -- they propose a $\rho$-$\kappa$ robustness guarantee, ensuring that with confidence higher than $\kappa$, with high probability $x$ has a robust radius above $\rho$. This work is conceptually close to ours in that it also trades pointwise guarantee for a statistically meaningful marginal guarantee. The distinction is that we do not aim to certify the global robustness of the classifier itself, and we do not offer a guaranteed mapping between confidence and robust radii. While we use a simple thresholded proxy as our example, one can define more complex functions and adapt it to our framework for free; additionally, we aim to replace the expensive certificate, with a cheaper function and not to state a confidence-radii front for the classifier. Our proxy is allowed to accept many inputs
cheaply, while its probability of accepting a genuinely non-robust input is controlled over the future data distribution. Intuitively, the distinction is analogous to the difference between conformal prediction and post-hoc calibration methods such as Platt scaling: PAG constructs a distributional guarantee over robustness events, whereas our method calibrates a proxy certificate to imitate an expensive certificate under a controlled marginal false-approval risk.

\paragraph{Conformal prediction and conformal risk control.}
Conformal prediction provides distribution-free finite-sample guarantees under exchangeability
\citep{Vovk2005AlgorithmicLI,Angelopoulos2021AGI,angelopoulos2024theoretical}. In classification,
set-valued predictors with valid coverage were studied by \citet{Sadinle2018LeastAS} and
\citet{Romano2020ClassificationWV}, and the same idea has since been used as a general-purpose
uncertainty quantification tool. Extensions beyond exact exchangeability, such as the weighted and
approximate guarantees studied by \citet{Barber2022ConformalPB}, are relevant whenever deployment
differs from calibration. Conformal risk control (CRC) extends the conformal guarantee from
miscoverage to general monotone risks \citep{angelopoulos2022conformal}. Instead of requiring the loss
to be the indicator of a missed label, CRC calibrates a parameter so that the expected value of a
user-chosen risk is below a target level. This is the main statistical primitive in our paper: the event
we need to control is not ordinary miscoverage, but false approval by a proxy certificate. Related
risk-control ideas also appear in risk-controlling prediction sets \citep{bates2021distribution}, in
the Learn then Test framework for finite-sample control of more general risks
\citep{angelopoulos2025learn}, and in selective prediction where a model learns when to abstain
\citep{geifman2017selective,geifman2019selectivenet}.
Our use of CRC differs from standard selective classification because the controlled failure event is
defined relative to adversarial robustness, and the test point may be a perturbation of an exchangeable
clean point rather than itself exchangeable with the calibration data.

\paragraph{Robust conformal prediction.}
Standard conformal guarantees do not directly survive adversarial test-time perturbations, since the
perturbed test input is no longer exchangeable with the clean calibration data. A common strategy is to
replace the conformity score by a worst-case score over the perturbation set. \citet{gendler2021adversarially}
combine conformal prediction with randomized smoothing in this way, obtaining adversarially robust
coverage under $\ell_2$ perturbations. \citet{Ghosh2023ProbabilisticallyRC} propose a probabilistic
variant that controls coverage for most perturbations rather than the worst case, improving the
efficiency-robustness trade-off. \citet{zargarbashirobust} derive tighter robust conformal sets by
bounding the worst-case change in conformity scores, covering both evasion and poisoning perturbations.
\citet{jeary2024verifiably} use verifiers to obtain robust conformal guarantees, and
\citet{massena2025efficient} pursue efficient robust conformal prediction through Lipschitz-bounded
networks. In a smoothing-based line, \citet{yan2024provably} revisit randomized-smoothed conformal prediction, identifying practical validity
issues and improving set efficiency. The closest recent lines are \citet{Zargarbashi2025OneSI}, who show
that robust conformal prediction can be obtained from a single randomly augmented sample by certifying
the conformal procedure itself, and \citet{anonymous2024robust}, who use a single binary certificate for
robust conformal prediction. Our work can be viewed as moving this single-sample idea from prediction
sets to proxy certification. We do not construct a set that covers the true label; instead, we calibrate
a proxy decision rule so that approving an unsafe perturbed input is rare.
More precisely, robust conformal risk control \citep{angelopoulos2022conformal} requires the worst-case loss over the perturbation set for every calibration point, which is itself a certification problem, and single-sample robust conformal prediction \citep{Zargarbashi2025OneSI} controls the coverage of a prediction set. Relative to these, the new ingredients of our work are: (i) the anchored false-approval risk (\autoref{thrm:proxy-cert}), defined on the pair of received point and latent clean anchor, which lets a sound certificate evaluated only at clean calibration points serve as the calibration signal, while the guarantee holds at the perturbed test point; and (ii) adapting the single-sample argument from coverage to general bounded monotone risks, which requires the concavity of $\upperval$ (\autoref{lemma:concave}).


\newpage
\section*{NeurIPS Paper Checklist}

\begin{enumerate}

\item {\bf Claims}
    \item[] Question: Do the main claims made in the abstract and introduction accurately reflect the paper's contributions and scope?
    \item[] Answer: \answerYes{}
    \item[] Justification: The abstract and introduction clearly describe the proposed framework, the conformal risk control guarantee (bounded marginal false-approval rate), both the standalone and fallback certificate use cases, and the empirical scope (vision transformers and language models). Aspirational goals are distinguished from proven guarantees.

\item {\bf Limitations}
    \item[] Question: Does the paper discuss the limitations of the work performed by the authors?
    \item[] Answer: \answerYes{}
    \item[] Justification: The paper includes a dedicated limitations paragraph (\emph{Fundamental limitations} in \autoref{sec:framework}) that discusses the gap between marginal and instance-wise guarantees, the dependence on proxy quality, and the exchangeability assumption underlying the conformal risk control framework.

\item {\bf Theory assumptions and proofs}
    \item[] Question: For each theoretical result, does the paper provide the full set of assumptions and a complete (and correct) proof?
    \item[] Answer: \answerYes{}
    \item[] Justification: All theoretical results are stated with explicit assumptions (e.g., exchangeability of the calibration set). Complete proofs are provided in the appendix, with proof sketches in the main text to convey intuition.

\item {\bf Experimental result reproducibility}
    \item[] Question: Does the paper fully disclose all the information needed to reproduce the main experimental results of the paper to the extent that it affects the main claims and/or conclusions of the paper (regardless of whether the code and data are provided or not)?
    \item[] Answer: \answerYes{}
    \item[] Justification: The calibration procedure, proxy tuning via conformal risk control, evaluation protocol, and all hyperparameter choices are described in sufficient detail in the paper and appendix. Code is also released at \url{https://github.com/soroushzargar/ProxC}.

\item {\bf Open access to data and code}
    \item[] Question: Does the paper provide open access to the data and code, with sufficient instructions to faithfully reproduce the main experimental results, as described in supplemental material?
    \item[] Answer: \answerYes{}
    \item[] Justification: The code is publicly available at \url{https://github.com/soroushzargar/ProxC}. All experiments rely on publicly available pretrained models and standard benchmarks.

\item {\bf Experimental setting/details}
    \item[] Question: Does the paper specify all the training and test details (e.g., data splits, hyperparameters, how they were chosen, type of optimizer) necessary to understand the results?
    \item[] Answer: \answerYes{}
    \item[] Justification: The paper specifies the calibration set sizes, risk level $\alpha$, proxy choices, and evaluation metrics. Full details are provided in the appendix.

\item {\bf Experiment statistical significance}
    \item[] Question: Does the paper report error bars suitably and correctly defined or other appropriate information about the statistical significance of the experiments?
    \item[] Answer: \answerYes{}
    \item[] Justification: Variance is reported for key experiments. Variance is generally very small, which is expected given the theoretical guarantees of the conformal risk control framework, and this is noted in the paper.

\item {\bf Experiments compute resources}
    \item[] Question: For each experiment, does the paper provide sufficient information on the computer resources (type of compute workers, memory, time of execution) needed to reproduce the experiments?
    \item[] Answer: \answerYes{}
    \item[] Justification: Formal hardware specifications (GPU type: NVIDIA H100 and A100) are reported in \autoref{sec:expr-appendix}.

\item {\bf Code of ethics}
    \item[] Question: Does the research conducted in the paper conform, in every respect, with the NeurIPS Code of Ethics \url{https://neurips.cc/public/EthicsGuidelines}?
    \item[] Answer: \answerYes{}
    \item[] Justification: The research is foundational methodology for robustness certification and conforms with the NeurIPS Code of Ethics. No human subjects, sensitive data, or dual-use concerns are involved.

\item {\bf Broader impacts}
    \item[] Question: Does the paper discuss both potential positive societal impacts and negative societal impacts of the work performed?
    \item[] Answer: \answerNA{}
    \item[] Justification: This is foundational methodology research on robustness certification without a direct path to specific deployment applications. The work improves the efficiency of existing safety tools and does not introduce new capabilities with foreseeable negative societal impacts.

\item {\bf Safeguards}
    \item[] Question: Does the paper describe safeguards that have been put in place for responsible release of data or models that have a high risk for misuse (e.g., pre-trained language models, image generators, or scraped datasets)?
    \item[] Answer: \answerNA{}
    \item[] Justification: The paper does not release new models or datasets. The released code implements a post-hoc certification method applied to existing publicly available models, posing no elevated misuse risk.

\item {\bf Licenses for existing assets}
    \item[] Question: Are the creators or original owners of assets (e.g., code, data, models), used in the paper, properly credited and are the license and terms of use explicitly mentioned and properly respected?
    \item[] Answer: \answerYes{}
    \item[] Justification: All pretrained models and datasets used in the experiments are cited with references to their original papers. Their usage is consistent with their intended research purposes.

\item {\bf New assets}
    \item[] Question: Are new assets introduced in the paper well documented and is the documentation provided alongside the assets?
    \item[] Answer: \answerYes{}
    \item[] Justification: The code implementing the proposed proxy certificate framework is released at \url{https://github.com/soroushzargar/ProxC} with documentation sufficient to reproduce the experiments.

\item {\bf Crowdsourcing and research with human subjects}
    \item[] Question: For crowdsourcing experiments and research with human subjects, does the paper include the full text of instructions given to participants and screenshots, if applicable, as well as details about compensation (if any)?
    \item[] Answer: \answerNA{}
    \item[] Justification: The paper does not involve crowdsourcing or research with human subjects.

\item {\bf Institutional review board (IRB) approvals or equivalent for research with human subjects}
    \item[] Question: Does the paper describe potential risks incurred by study participants, whether such risks were disclosed to the subjects, and whether Institutional Review Board (IRB) approvals (or an equivalent approval/review based on the requirements of your country or institution) were obtained?
    \item[] Answer: \answerNA{}
    \item[] Justification: The paper does not involve human subjects research and requires no IRB review.

\item {\bf Declaration of LLM usage}
    \item[] Question: Does the paper describe the usage of LLMs if it is an important, original, or non-standard component of the core methods in this research? Note that if the LLM is used only for writing, editing, or formatting purposes and does \emph{not} impact the core methodology, scientific rigor, or originality of the research, declaration is not required.
    \item[] Answer: \answerNA{}
    \item[] Justification: LLMs appear in the paper only as one of the evaluated classifier architectures (i.e., as objects to be certified), not as a component of the proposed method. LLMs were used solely for writing assistance and proofreading, which does not require declaration per NeurIPS policy.

\end{enumerate}


\end{document}